\documentclass[sigconf, 9pt]{acmart}
\setcitestyle{numbers,sort&compress,mcite,merge}
\renewcommand\footnotetextcopyrightpermission[1]{} 
\AtBeginDocument{%
  \providecommand\BibTeX{{%
    \normalfont B\kern-0.5em{\scshape i\kern-0.25em b}\kern-0.8em\TeX}}}

\usepackage{amsfonts}
\usepackage{mathtools}
\usepackage{graphicx}
\usepackage{subfigure}
\usepackage{enumerate}
\usepackage{amsmath}
\usepackage{color}
\usepackage{amsthm}
\usepackage{algorithm}
\usepackage{algpseudocode}
\usepackage{stfloats}
\usepackage{multirow}
\usepackage{bbm}
\usepackage{array}
\newcolumntype{C}[1]{>{\centering\arraybackslash}p{#1}}

\newcommand\EP{\texttt{E$^3$Pose}}

\begin{document}

\title{E$^3$Pose: Energy-Efficient Edge-assisted Multi-camera System for Multi-human 3D Pose Estimation}

\author{Letian Zhang}
\email{lxz437@miami.edu}
\affiliation{%
	\institution{University of Miami}
	\city{Coral Gables}
	\state{Florida}
	\country{USA}
	\postcode{33146}
}

\author{Jie Xu}
\email{jiexu@miami.edu}
\affiliation{%
	\institution{University of Miami}
	\city{Coral Gables}
	\state{Florida}
	\country{USA}
	\postcode{33146}
}

\begin{abstract}
Multi-human 3D pose estimation plays a key role in establishing a seamless connection between the real world and the virtual world. Recent efforts adopted a two-stage framework that first builds 2D pose estimations in multiple camera views from different perspectives and then synthesizes them into 3D poses. However, the focus has largely been on developing new computer vision algorithms on the offline video datasets without much consideration on the energy constraints in real-world systems with flexibly-deployed and battery-powered cameras. In this paper, we propose an energy-efficient edge-assisted multiple-camera system, dubbed \EP, for real-time multi-human 3D pose estimation, based on the key idea of adaptive camera selection. Instead of always employing all available cameras to perform 2D pose estimations as in the existing works, \EP~ selects only a subset of cameras depending on their camera view qualities in terms of occlusion and energy states in an adaptive manner, thereby reducing the energy consumption (which translates to extended battery lifetime) and improving the estimation accuracy. To achieve this goal, \EP~ incorporates an attention-based LSTM to predict the occlusion information of each camera view and guide camera selection before cameras are selected to process the images of a scene, and runs a camera selection algorithm based on the Lyapunov optimization framework to make long-term adaptive selection decisions. We build a prototype of \EP~ on a 5-camera testbed, demonstrate its feasibility and evaluate its performance. Our results show that a significant energy saving (up to 31.21\%) can be achieved while maintaining a high 3D pose estimation accuracy comparable to state-of-the-art methods. 
\end{abstract}


\maketitle
\pagestyle{plain}

\section{Introduction}
Multi-human 3D pose estimation is an important yet challenging computer vision problem that has a wide range of applications such as action recognition \cite{luvizon2020multi}, sports analysis \cite{bridgeman2019multi} and human-computer interaction \cite{zheng2022multi}. It is also perceived as a key technology to enable the seamless and immersive interaction between the real world and the virtual world in the so-called ``Metaverse'' \cite{wang2022best}. Despite the rapid development of deep learning methods, monocular-view-based approaches \cite{sun2018integral,cheng2019occlusion,zhang2020inference} still suffer from large errors in practice due to occlusion, motion blur and lack of the absolute world coordinate in the single camera setup. Recent efforts have been shifted to studying multi-view approaches \cite{dong2019fast, chen2020cross, lin2021multi, tanke2019iterative} where the 3D poses are constructed from multiple camera views. Typically, a two-stage framework is employed where 2D poses are first estimated from the individual camera views and then 3D poses are reconstructed based on these 2D poses. Although much progress has been made, the majority of existing works study the problem from a pure computer vision perspective, neglecting the system constraints in a real-world deployment scenario for real-time multi-human 3D estimation. 

The two-stage framework of multi-view 3D pose estimation naturally fits into an edge-assisted multi-camera system \cite{simon2021}. Specifically, distributed smart cameras perform multi-human 2D pose estimation on their individual monocular views, which are not only able to capture high-resolution video data but also equipped with hardware accelerators to execute deep learning-based video processing. The 2D pose estimation results are then transmitted wirelessly to a relatively powerful edge server for the 3D pose reconstruction. Compared to sending the raw video frames from the cameras to the edge server, sending only the 2D pose estimation results can significantly reduce the transmitted data size of the cameras and overcome the wireless bottleneck. On the flip side, however, moving the workload of 2D pose estimation to the cameras demands much higher energy usage of the cameras. For instance, running lightweight deep neural network (DNN)-based multi-human 2D pose estimation requires twice of the power consumption in the standby mode on NVIDIA Jetson Xavier NX \cite{nvidianx}, a popular DL-capable embedded system. Therefore, excessive energy consumption may become a major obstacle to the flexible deployment of edge-assisted 3D pose estimation system with battery-powered cameras. 

In this paper, we present the design, implementation and evaluation of an energy-efficient edge-assisted multi-camera system for multi-human 3D pose estimation, dubbed \EP. At the core of \EP~  is a camera selection scheme that adaptively selects a subset of cameras to perform the 3D pose estimation task depending on the cameras' energy states and their view qualities (i.e., occlusions). Obviously, performing the 3D pose estimation using only a subset of cameras allows some cameras to enter a power-saving mode, thereby extending the battery lifetime of the cameras and the overall system. Less obviously, a careful selection of the subset of cameras to participate in 3D pose estimation can still obtain a comparable, sometimes even higher, estimation accuracy compared to the default full participation case. This is because, in theory, 2D poses from just a few clear camera views are sufficient to synthesize the final 3D poses and, in practice, cameras with clear views vary over time depending on the scene. To realize the function of \EP, two key challenges must be addressed as follows.

\textbf{How to predict the cameras' view qualities of a future scene at the selection time?} Because of the various delays in data processing and transmission, camera selection decisions must be made before a scene comes up. More critically, because the selection decision is made before the selected cameras perform 2D pose estimation on a scene, the unselected cameras have no way to evaluate their view qualities in terms of occlusion since they are not supposed to perform 2D estimation. Thus, for the purpose of camera selection, the cameras' view qualities of a future scene of interest must be \textit{predicted}. To cope with this challenge, we design an attention-based LSTM network to predict 3D human poses in a future scene (at a future time) based on the 3D human pose information in the current scene (at the current time). The predicted 3D pose results are then projected to 2D views for individual cameras to calculate the occlusion value of each camera. Although the predicted 3D pose results are not (and need not be) perfectly accurate, they provide sufficiently valuable occlusion information for camera selection. 

\textbf{How to select the cameras to achieve the long-term energy efficiency while considering their view qualities?} To extend the battery lifetime of the overall system, 2D pose estimation workload must be balanced among the distributed cameras without creating processing hotspots on just a few cameras. To this end, we develop a camera scheduler deployed on the edge server to make camera selection decisions in an online fashion by leveraging the Lyapunov optimization framework. When making the selection decision, the scheduler takes into consideration of both the current battery state and the predicted view quality of the cameras to make a trade-off between the long-term energy consumption and the 3D pose estimation accuracy. The edge server acts as a convenient anchor point to make the selection decision without putting extra work on the energy-limited cameras. 

We implement \EP~on a testbed, where 5 smart camera devices connect wirelessly to an edge server. Evaluations on existing real-world datasets and live video streams on the testbed were conducted, which show that \EP~is able to achieve a significant energy consumption reduction while maintaining a high multi-human 3D pose estimation accuracy. 

\section{Motivational Experiments}
\begin{figure}[tt]
	\centering
	\includegraphics[width=0.95\linewidth]{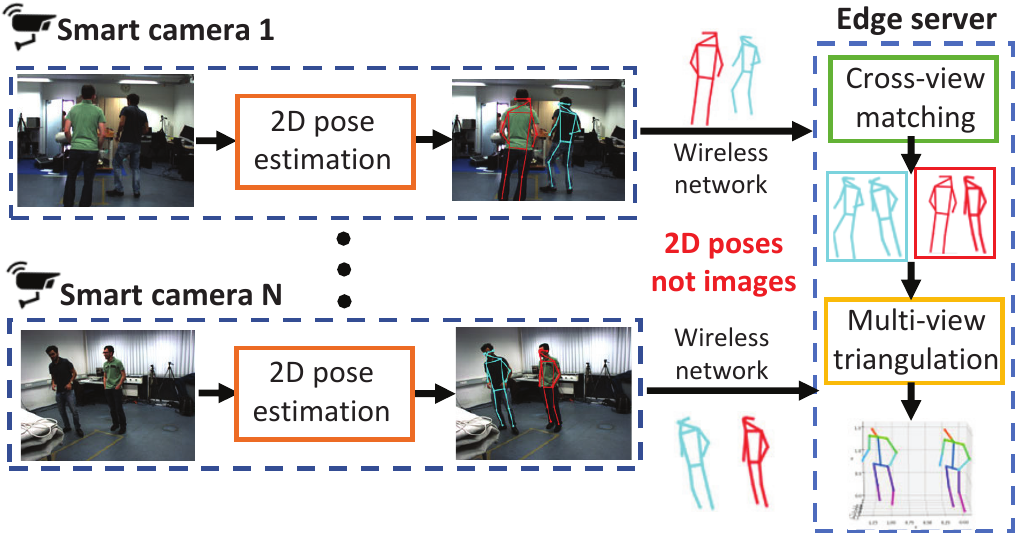}
    \setlength{\abovecaptionskip}{0.cm}
	\caption{Illustration of edge-assisted 3D pose estimation. Distributed cameras first perform 2D pose estimation on the captured image and then the edge server synthesizes 3D poses based on the estimated 2D poses.}
	\label{fig:edgeAssistedPose}
	\vspace{-20 pt}
\end{figure}

The generic setup of an edge-assisted multi-camera system for multi-human 3D pose estimation is shown in Figure \ref{fig:edgeAssistedPose}. First, distributed smart cameras perform 2D pose estimation on their captured video frames. The 2D human pose results are then streamed over the wireless network to a central edge server, where data association, cross-view matching, multi-view triangulation and post-processing are performed to synthesize the 2D pose results into 3D skeletons. We consider an energy-limited (e.g., battery-powered) multi-camera system that allows flexible deployment and hence energy efficiency is a key design objective for usability. Although energy consumption is less concerned for plug-in cameras, the stable power supply may not always be available and wiring represents a major obstacle for outdoor deployment. In fact, increasingly many outdoor and in-field camera systems are powered by battery and/or energy harvesting devices (e.g., solar panels) \cite{wirelessCCTV,sunSurveillance}. In what follows, we provide experimental evidence to demonstrate the opportunities for designing an energy-efficient edge-assisted 3D pose estimation system.

\begin{figure}[tt]
	\centering  
	\subfigure[NVIDIA Jetson Xavier NX]{
		\label{fig:powerMotivationXavier}
		\includegraphics[width=0.48\linewidth]{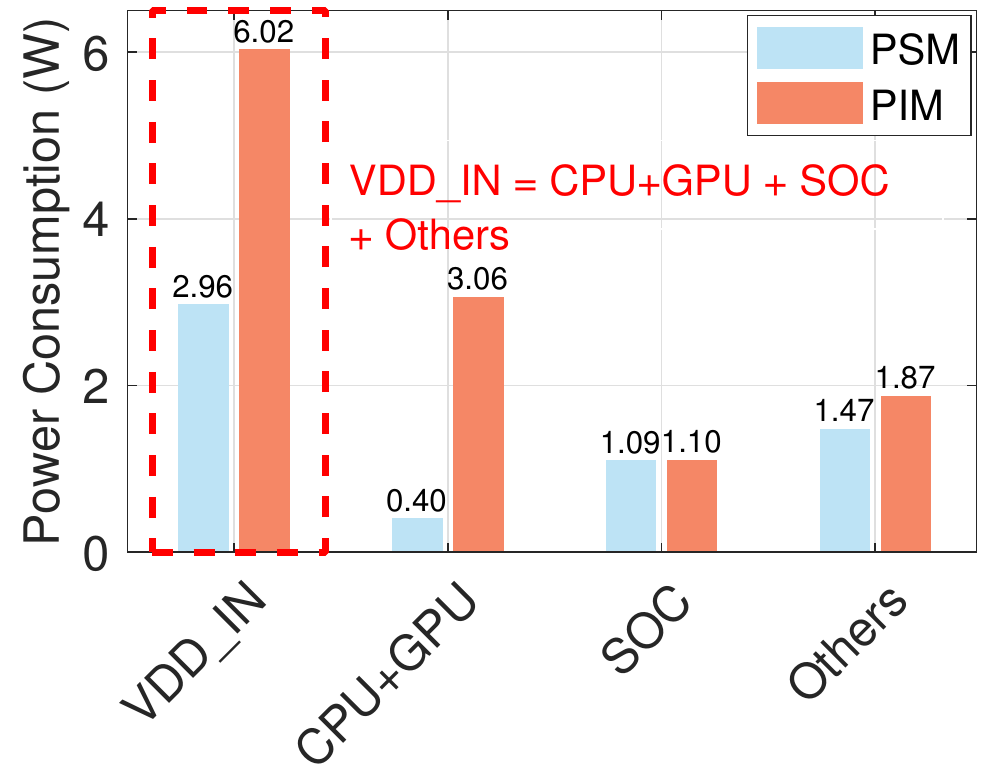}}
	\subfigure[NVIDIA Jetson TX2]{
		\label{fig:powerMotivationTX2}
		\includegraphics[width=0.48\linewidth]{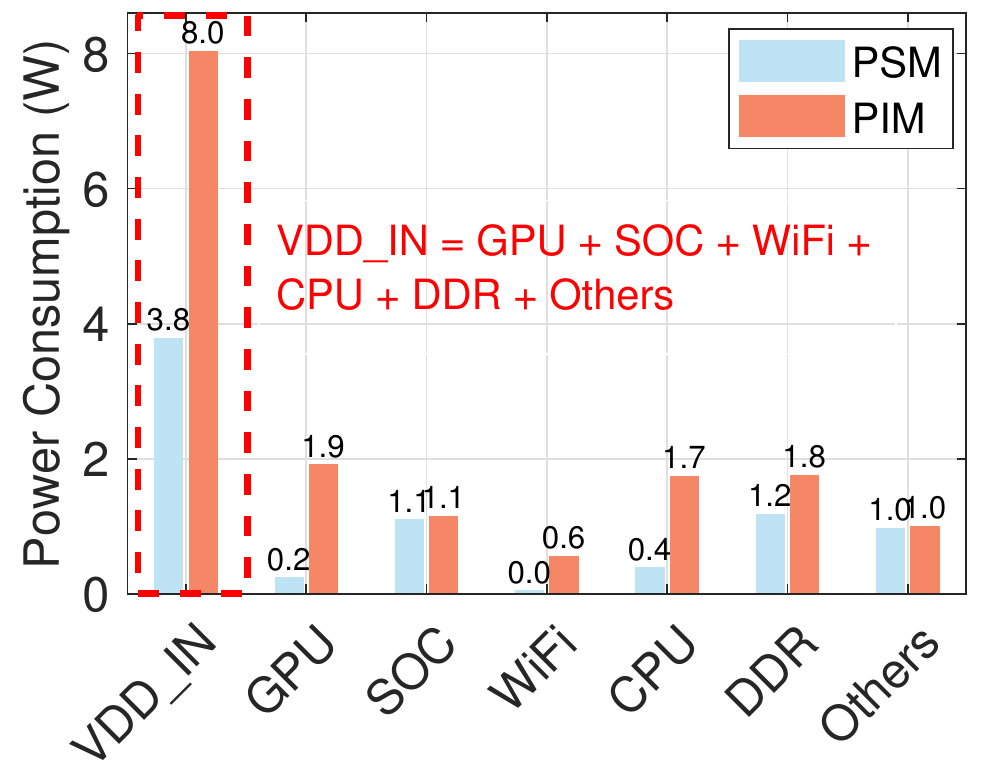}}
	\setlength{\abovecaptionskip}{0.cm}
	\caption{Power consumption of cameras in different modes. (The power consumption monitors of NVIDIA Jetson Xavier NX and NVIDIA Jetson TX2 give measurement results in different granularity.)}
	\label{fig:PowerConsumptionMotivation}
\end{figure}

\subsection{Power Consumption}
To understand how much benefits in terms of energy saving can be obtained by selecting only a subset of cameras to perform 3D pose estimation, we conduct experiments to measure the power consumption of smart camera devices in the Power-Intensive Mode (PIM) and the Power-Saving Mode (PSM). Specifically, these two modes are defined as follows: \textbf{Power-Intensive Mode}: the camera captures the video frames, runs a DNN to obtain the 2D human pose result on the local device and sends the 2D pose estimation result to the edge server via wireless. \textbf{Power-Saving Mode}: the camera captures the frames but does not perform 2D human pose estimation or communicate the result with the edge server.

Figure \ref{fig:PowerConsumptionMotivation} illustrates the power consumption profiles of two camera devices used in our testbed, one based on NVIDIA Jetson Xavier NX \cite{nvidianx} and the other based on NVIDIA Jetson TX2 \cite{nvidiatx2}, which are measured using the embedded power consumption monitors in the NVIDIA Jetson devices. In particular, Figure \ref{fig:powerMotivationXavier} shows the total power consumption (VDD\_IN) of Xavier NX in two different modes, and the breakdown in terms of processing units (CPU + GPU), base system (SOC) and other components (Others). TX2 offers a finer breakdown information of the power consumption and hence, Figure \ref{fig:powerMotivationTX2} shows the total power consumption (VDD\_IN) and the power consumption of GPU, CPU, WiFi, DDR, SOC and Others. As can be seen, on both devices, PSM reduces power consumption by roughly half compared to PIM, where the power saving mainly comes from the processing units and the wireless data transmission. These results suggest a great potential of letting some cameras enter the PSM to save energy, thereby extending the battery lifetime of the overall system. Note that the switching cost (mostly delay) between the two modes is negligible as the DNN for 2D human pose estimation is still cached in memory even in the PSM.

\subsection{3D Pose Estimation Accuracy} \label{sec:accuracy_3dfusion}
Multi-human 3D pose estimation solutions \cite{belagiannis20143d, belagiannis2014multiple, dong2019fast, chen2020cross, lin2021multi, tanke2019iterative, qiu2019cross, remelli2020lightweight} in the computer vision literature focus on how to improve the estimation accuracy with a \textit{given} set of camera views. As a straightforward application of these solutions to an edge-assisted multi-camera system, all smart cameras by default perform 2D human pose estimation and send the results to the edge server \cite{simon2021}. In the previous subsection, we showed that using only a subset of cameras can potentially achieve a significant energy consumption reduction, but what is unclear is whether such an energy saving comes at the cost of a lowered estimation accuracy. In this set of experiments, we investigate the impact of activating different subsets of cameras on the 3D pose estimation accuracy. We find that it is not always necessary to use all cameras' 2D pose results to obtain an accurate 3D pose estimation result. In fact, a judiciously selected subset of cameras can even outperform the entire set of cameras in many cases. This is because errors in the 2D estimation results of cameras with severe occlusions can bring down the accuracy of the 3D pose estimation when these cameras are included in the 3D pose synthesis.

\begin{figure*}[tt]
	\centering
	\includegraphics[width=\linewidth]{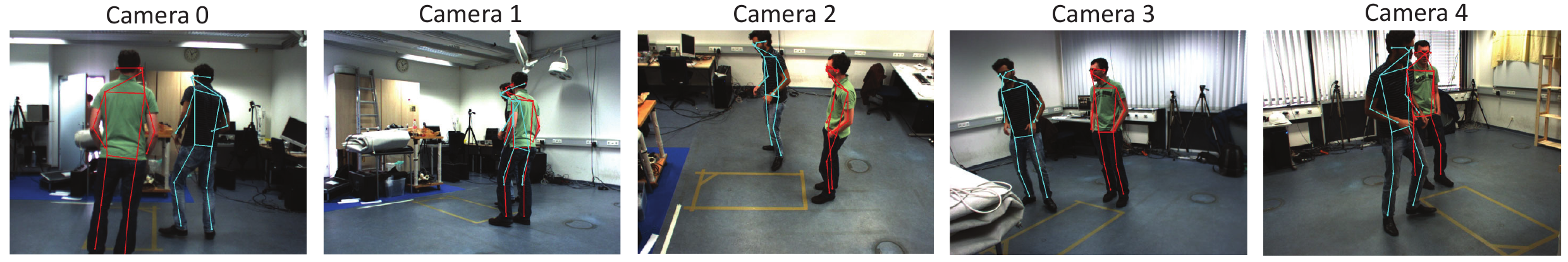}
    \vspace{-20 pt}
	\caption{One representative scene in the Shelf dataset and the estimated 2D poses in five camera views. Due to occlusion, the estimation results in Camera 1 and Camera 4 have larger errors. }
	\label{fig:multiview2DPoses}
	\vspace{-10 pt}
\end{figure*}

\begin{figure}[tt] 
	\centering  
	\subfigure[Camera 0, 1, 2, 3, 4]{
		\label{fig:fiveCameras}
		\includegraphics[width=0.42\linewidth]{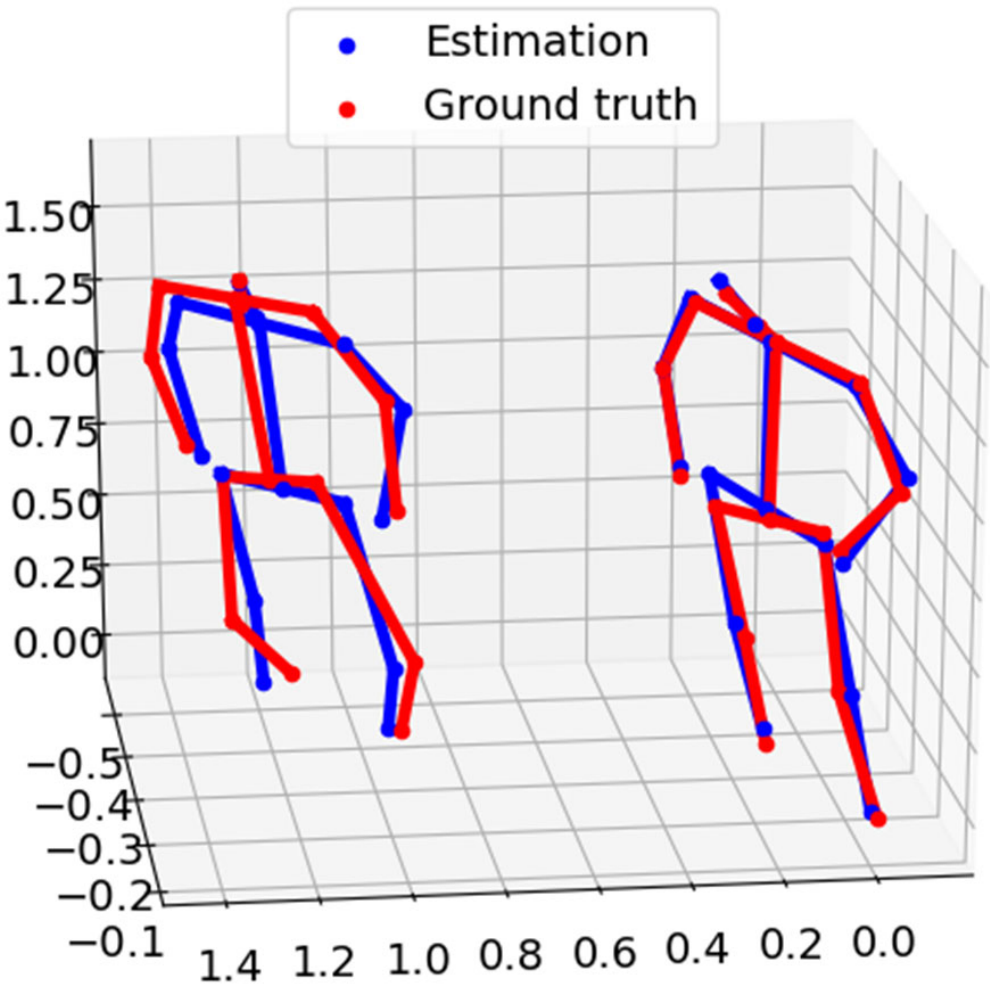}}
	\subfigure[Camera 0, 2, 3]{
		\label{fig:threeCameras}
		\includegraphics[width=0.42\linewidth]{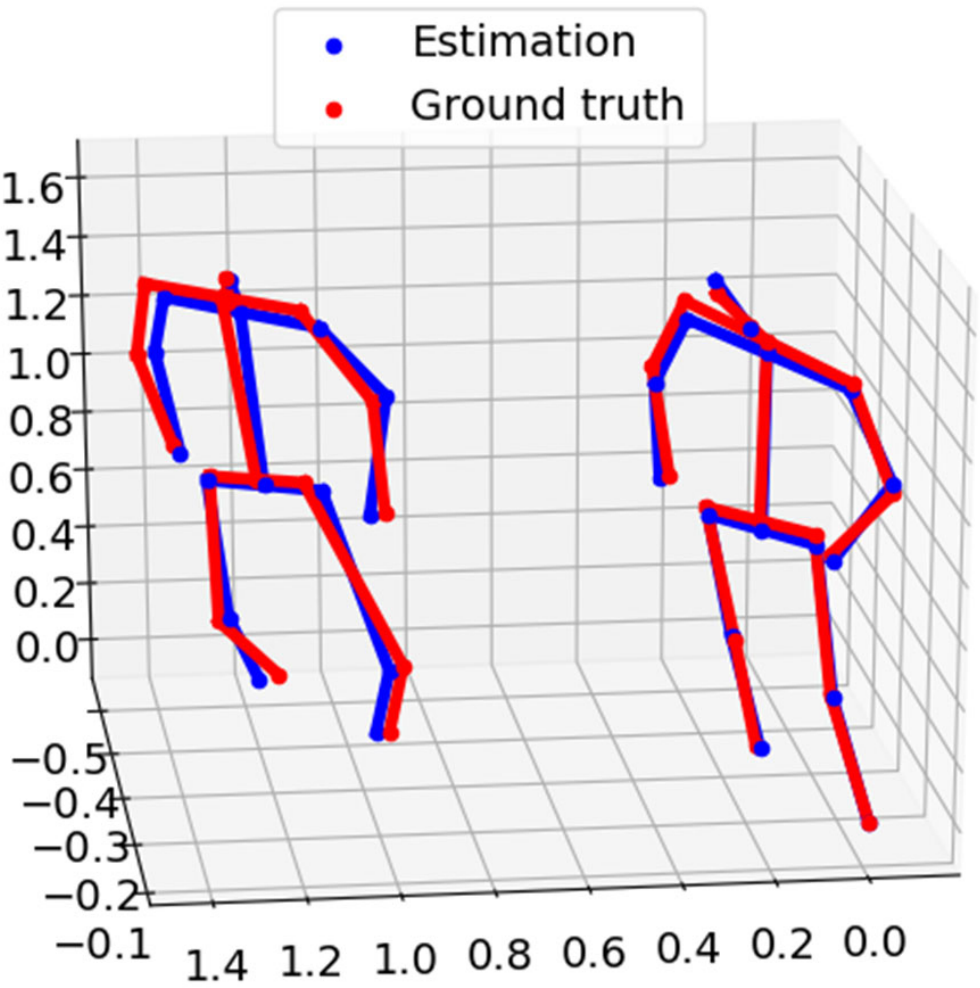}}
	\vspace{-15 pt}
	\caption{3D pose estimation results obtained on different combinations of cameras. The estimation error is smaller with only a subset of cameras in this specific case. }
	\label{fig:cameraComparison}
\end{figure}

To support the above findings, we illustrate the 3D pose estimation results on the Shelf dataset \cite{belagiannis20143d}, which contains 3200 scenes from 5 camera views. Figure \ref{fig:multiview2DPoses} shows the estimated 2D poses in all 5 camera views for a representative scene. As can be seen, depending on the relative positions and postures of the human objects in the scene, different camera views manifest different occlusion patterns. In particular, since the occlusions in Camera 1 and Camera 4 are more severe, the respective 2D pose estimation results also contain larger errors. Furthermore, Figure \ref{fig:cameraComparison} shows that the 3D poses reconstructed by using only the 2D results of Cameras 0, 2, 3 incur a much smaller error than those reconstructed by using all Cameras' 2D results through a visual inspection. 

\begin{figure}[tt]
	\centering  
	\subfigure[]{
		\label{fig:mpjpePcpMotivation}
		\includegraphics[width=0.48\linewidth]{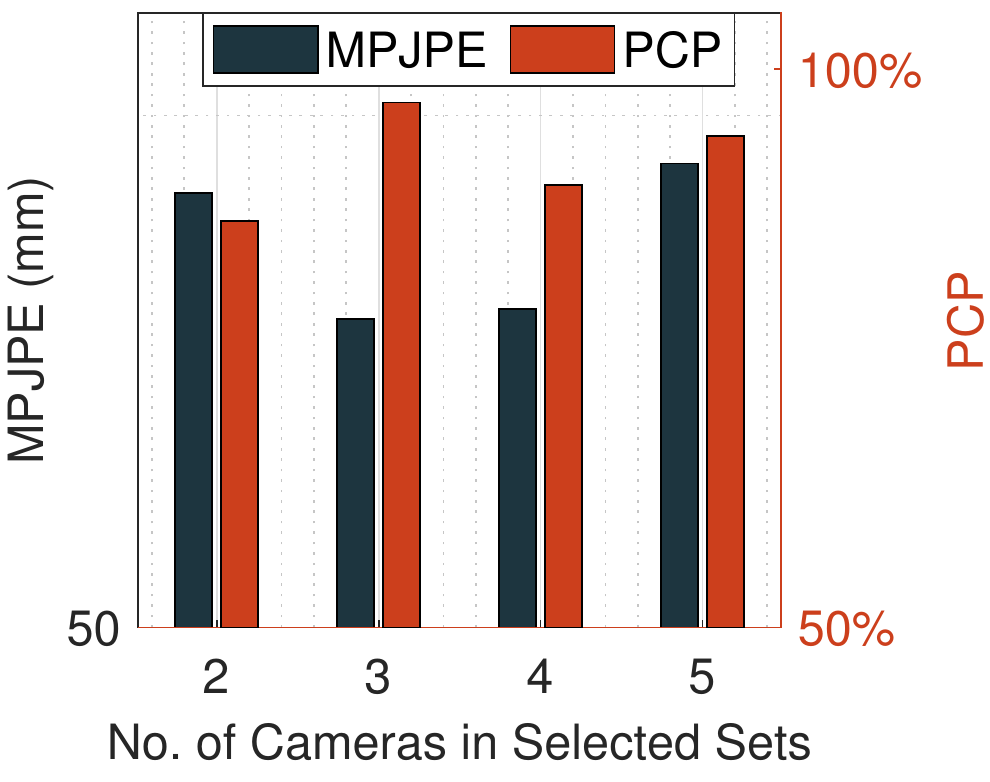}}
	\subfigure[]{
		\label{fig:pcpCameraCombination}
		\includegraphics[width=0.48\linewidth]{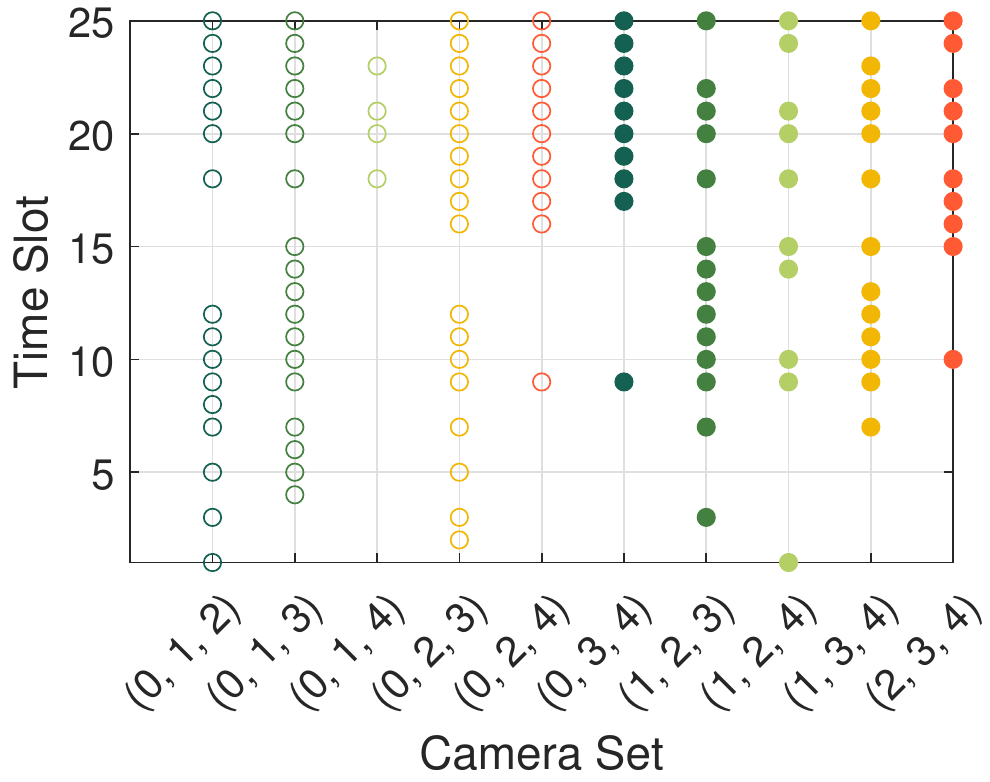}}
	\vspace{-15 pt}
	\caption{Accuracy performance of different camera sets. (a) 3-camera subsets achieve the highest PCP and the lowest MPJPE. (b) The best subset(s) of 3 cameras that achieve the highest PCP varies across scenes (time slots). }
	\label{fig:accuracyMotivation}
\end{figure}

Figure \ref{fig:accuracyMotivation} further provides numerical results to demonstrate that a subset of cameras often outperforms the entire set of cameras for 3D pose estimation. We measure the 3D pose estimation performance by both the Mean Per Joint Position Error (MPJPE) and the Percentage of Correct Parts (PCP) between the estimated 3D pose and the ground truth,  following the same evaluation protocol in the literature \cite{tanke2019iterative,qiu2019cross, remelli2020lightweight,chen2020cross,dong2019fast}. For each scene, we test all camera combinations, and report the average best estimation results for different numbers of cameras in the selected subset in Figure \ref{fig:mpjpePcpMotivation}. We observe that the average estimation performance in terms of both MPJPE and PCP is the best when a subset of 3 cameras are selected to process the scene, rather than using all 5 cameras. 

Note, however, that the best subset of 3 cameras are not static but rather changes over time depending on the scene. In Figure \ref{fig:pcpCameraCombination}, we show that the best combination (i.e., the highest PCP) of a 3-camera subset varies in different scenes over time. This suggests that although deploying only 3 cameras reduces the total energy consumption than deploying 5 cameras, it does not extend the battery lifetime of the system (because every camera is in the PIM all the time) or guarantee an improved the 3D pose estimation accuracy. On the contrary, we advocate a design where a relatively large number of cameras are deployed (considering the increasingly affordable device cost and the deployment flexibility due to wireless transmission and battery power supply), but through \textit{adaptive} camera selection, the system battery lifetime can be extended and the 3D pose estimation accuracy can be improved.

\section{System Design}
\begin{figure}[tt]
	\centering
	\includegraphics[width=0.95\linewidth]{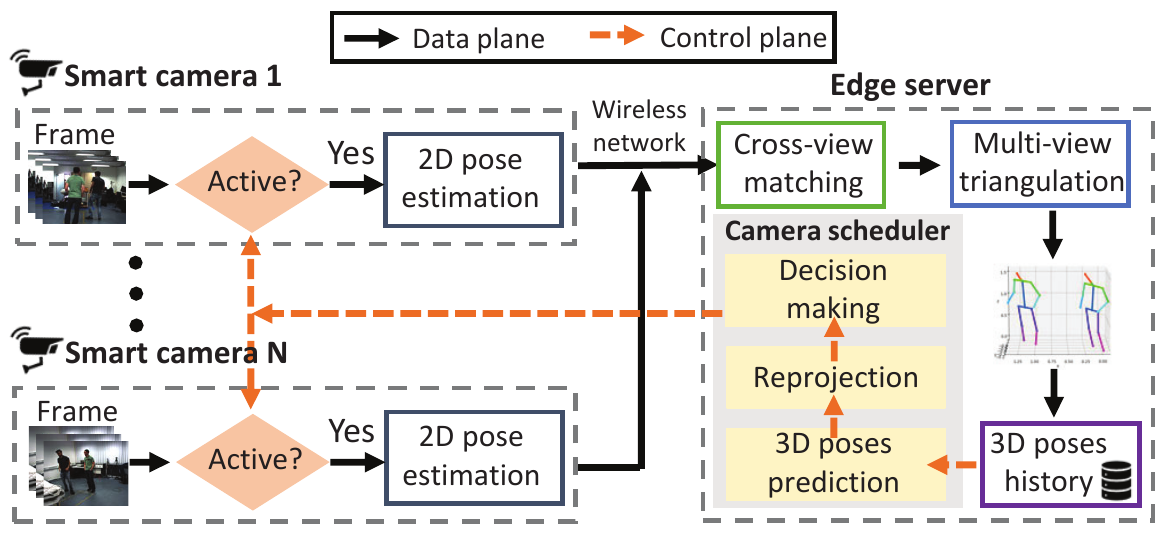}
    \setlength{\abovecaptionskip}{0.cm}
	\caption{The system architecture and workflow of $E^3$Pose.}
	\label{fig:systemArchitecture}
\end{figure}

In this section, we design \EP, an energy-efficient edge-assisted multi-camera system that enables real-time multi-human 3D pose estimation based on the core idea of adaptive camera selection. We consider a typical setup where the edge server is a relatively powerful machine that is powered by a stable power source (e.g., the power grid), while the cameras are lightweight and battery powered to enable flexible deployment. A typical number of cameras is 5 or 6, but more cameras can also be deployed as the cost of cameras becomes increasingly lower, in order to extend the system battery lifetime while achieving a high 3D pose estimation accuracy. Figure \ref{fig:systemArchitecture} shows the system architecture and workflow of \EP. In what follows, we first provide an overview of the data plane and control plane of $\EP$~ and then describe the design details. 

\textbf{Data Plane}. For each input frame captured by the camera, the camera either runs the DNN to estimate the 2D human poses if it is in the PIM (the \textit{Active} switch is Yes) or skips processing this frame if it is in the PSM (the \textit{Active} switch is No). The mode switch is controlled by the camera scheduler module residing at the edge server. Once the cameras in the PIM finish the 2D pose estimation, they send the results to the edge server via wireless for 3D pose fusion. Note that compared with the raw input images, the 2D poses are represented by just several pixel coordinates of human joints. Therefore, sending the 2D pose results to the edge server incurs a very small wireless transmission cost. Then, using the received 2D poses from the smart cameras in the PIM, the edge server matches the human poses across these 2D poses and synthesizes the matched multi-view 2D poses into 3D poses. The edge server also saves the 3D poses in each time slot for future user query.

\textbf{Control Plane}. The camera scheduler runs a prediction module
to predict the occlusion in each camera view of a scene in a future time slot $t'$. This is done by first predicting the 3D poses at $t'$ based on the 3D poses in the current time slot $t$ and the past time slots. The predicted 3D poses are then projected onto each camera's 2D view. Next, bounding boxes on the projected 2D poses are generated, which are used to calculate the occlusion of persons in each camera's view. Based on the calculated occlusion and the energy state of the cameras, the camera scheduler makes the control decision to select the subset of cameras (turn the \textit{Active} switch on/off at time $t'$). Note that we intentionally make the scheduling decisions ahead of time to ensure that the control signals can arrive at the cameras before time $t'$. Apparently, the time advance $t'-t$ must be larger than the total delay due to prediction computation and control signal transmission.

\textit{\textbf{Remark on Terminologies}}: We clarify the difference between ``3D Pose Estimation'' and ``3D Pose Prediction'' used in this paper. 3D Pose Estimation is the main function of \EP, which refers to using the 2D poses estimated by individual cameras on their captured images from different perspectives of the scene to estimate the 3D poses. As part of \EP~, 3D Pose Prediction module is designed to get a rough idea of the 3D pose in a future scene so that the occlusion information of camera views can be calculated for the future scene. However, since the purpose of 3D pose prediction is mere to calculate/predict occlusion, its accuracy does not need to be as high as 3D pose estimation.

\subsection{Data Plane}
The data plane in \EP~ has two main modules -- cross-view matching and multi-view triangulation. Cross-view matching collects the 2D poses from the cameras and determines which 2D poses from different cameras are indeed the same person. Then, based on the matched 2D poses, multi-view triangulation fuses the 2D poses into the 3D poses. 

\subsubsection{\textbf{Cross-view matching}}
Consider an \EP~ system containing $N$ smart cameras with known projection matrices $P_n \in \mathbb{R}^{3 \times 4}, \forall n = 1, ..., N$, which can be easily obtained during system setup. We also assume that they have the same frame rate. For convenience, we consider a time-slotted system where in each time slot, a frame is captured by a camera. We assume that the clocks of smart cameras and the edge server are software synchronized and each 2D pose message includes a timestamp representing the capture time of the corresponding frame. Therefore, frames from cameras in the same time slot are image captures of the same scene at the same time from different perspectives. 

Before performing 3D pose synthesis, the detected 2D poses must be matched across views so that 2D poses in all views belonging to the same person are found. We use an efficient iterative greedy matching algorithm proposed in \cite{tanke2019iterative} to associate the 2D poses across camera views. The key idea of this method is to make sure that the associated 2D poses from different views satisfy the epipolar constraint, i.e., a human body joint in one view should lie on the epipolar line associated with its correspondence in another view. For the matching, we select a camera as the starting point and choose all 2D human poses in this camera view as person candidates. We then iterate over all the other cameras and match their 2D poses with the current list of person candidates in a greedy manner using the distance between the epipolar lines and the joint locations. 

\subsubsection{\textbf{Multi-view triangulation}}
Once cross-view matching is done, each human object is associated with a set of 2D poses. Consider a representative human object in a particular time slot, who is associated with a set of 2D poses $\mathbf{x}_n = \{x^j_n \in \mathbb{R}^2; j = 1, 2, ..., J\}, \forall n \in \mathcal{H}$ where $\mathcal{H}$ is the set of cameras that are selected and contain the human object in that particular time slot, and $J$ is the number of human joints that are used to indicate the human pose skeleton. Each point $x^j_n = [u^j_n, v^j_n]$ in the 2D pose $\mathbf{x}_n$ represents the 2D pixel coordinate of the $j$-th human body joint. These 2D poses will be used to synthesize the 3D pose of the human object, which is denoted by $\mathbf{X} = \{X^j \in \mathbb{R}^3; j = 1, 2, ..., J\}$. Specifically, the relation between $\{\mathbf{x}_n\}_{n \in \mathcal{H}}$ and $\mathbf{X}$ is as follows:
\begin{equation} \label{eq:triangulateFunc}
	(\omega^j \circ A^j) \tilde{X}^j = 0, ~~ j = 1, \dots, J
\end{equation}
\begin{equation}
	A^j = \begin{bmatrix}
		u_n^j p_{n, 3} - p_{n, 1} \\
		v_n^j p_{n, 3} - p_{n, 2} \\
	\end{bmatrix}_{n \in \mathcal{H}} \in \mathbb{R}^{2|\mathcal{H}| \times 4}
\end{equation}
\begin{equation}
	\omega^j = \left(\frac{c_n^j}{||(A^j)_n^1||}, \frac{c_n^j}{||(A^j)_n^2||}\right)_{n \in \mathcal{H}} \in \mathbb{R}^{2|\mathcal{H}|}
\end{equation}
where $\tilde{X}^j \in \mathbb{R}^4$ is the homogeneous coordinates of $X^j$, $p_{n, r}$ denotes the $r$-th row of the projection matrix $P_n \in \mathbb{R}^{3 \times 4}$, $c_n^j$ is heatmap confidence value of joint $j$ in camera $n$'s 2D human pose estimation, $(A^j)_n^1$ and $(A^j)_n^2$ are the rows of $A^j$ corresponding to camera $n$ and $\circ$ is the Hadamard product. When calculating $\omega^j$, $c_n^j$ is divided by the $L_2$-norm of the corresponding row of $A^j$ to compensate for the different image locations of the joints in each view. According to the Direct Linear Transform (DLT) algorithm \cite{hartley2003multiple}, if there are at least two views, Equation \eqref{eq:triangulateFunc} is overdetermined and can be solved by a singular value decomposition (SVD) on $\omega^j \circ A^j$, taking the unit singular vector corresponding to the smallest singular value of $\omega^j \circ A^j$ as solution for $\tilde{X}^j$. Finally, $\tilde{X}^j$ is divided by its fourth coordinate to obtain the 3D joint $X^j = \tilde{X}^j / (\tilde{X}^j)_4$.

\subsection{Control Plane}
At the core of the control plan is the camera scheduler that selects the subset of cameras to perform multi-human 3D pose estimation for each scene. Consider the scheduling problem for the scene in time slot $t$. We use Intersection over Union (IoU) of the bounding boxes, denoted by $\text{IoU}^t_n$ for camera $n$ in time slot $t$, to quantify the occlusion in camera $n$'s view. Originally, IoU is used to measure the overlapping area of the predicted bounding boxes and the ground truth in object detection. Here, we repurpose IoU to measure the overlapping area among the predicted bounding boxes of different human objects in the camera view. Specifically, let $R^t_{n, 1}, ..., R^t_{n, K^t_n}$ be the areas of the $K^t_n$ bounding boxes in camera $n$'s view in time slot $t$. Then, $\text{IoU}^t_n$ can be calculated as
\begin{equation} \label{eq:iouCalculation}
    \setlength{\abovedisplayskip}{3pt}
    \setlength{\belowdisplayskip}{3pt}
	\text{IoU}^t_n = \frac{1}{\binom{K^t_n}{2}}\sum_{k_1 = 1}^{K^t_n} \sum_{k_2 = k_1 + 1}^{K^t_n} \frac{R^t_{n, k_1} \cap R^t_{n, k_2}}{R^t_{n, k_1} \cup R^t_{n, k_2}}
\end{equation}
where $\binom{K^t_n}{2}$ is the number of 2-combinations from a given $K^t_n$ elements. Note that $\text{IoU}^t_n \in [0, 1]$ with a value 0 representing no occlusion and a value 1 representing complete overlapping. 

\begin{figure}[tt]
	\centering
	\includegraphics[width=\linewidth]{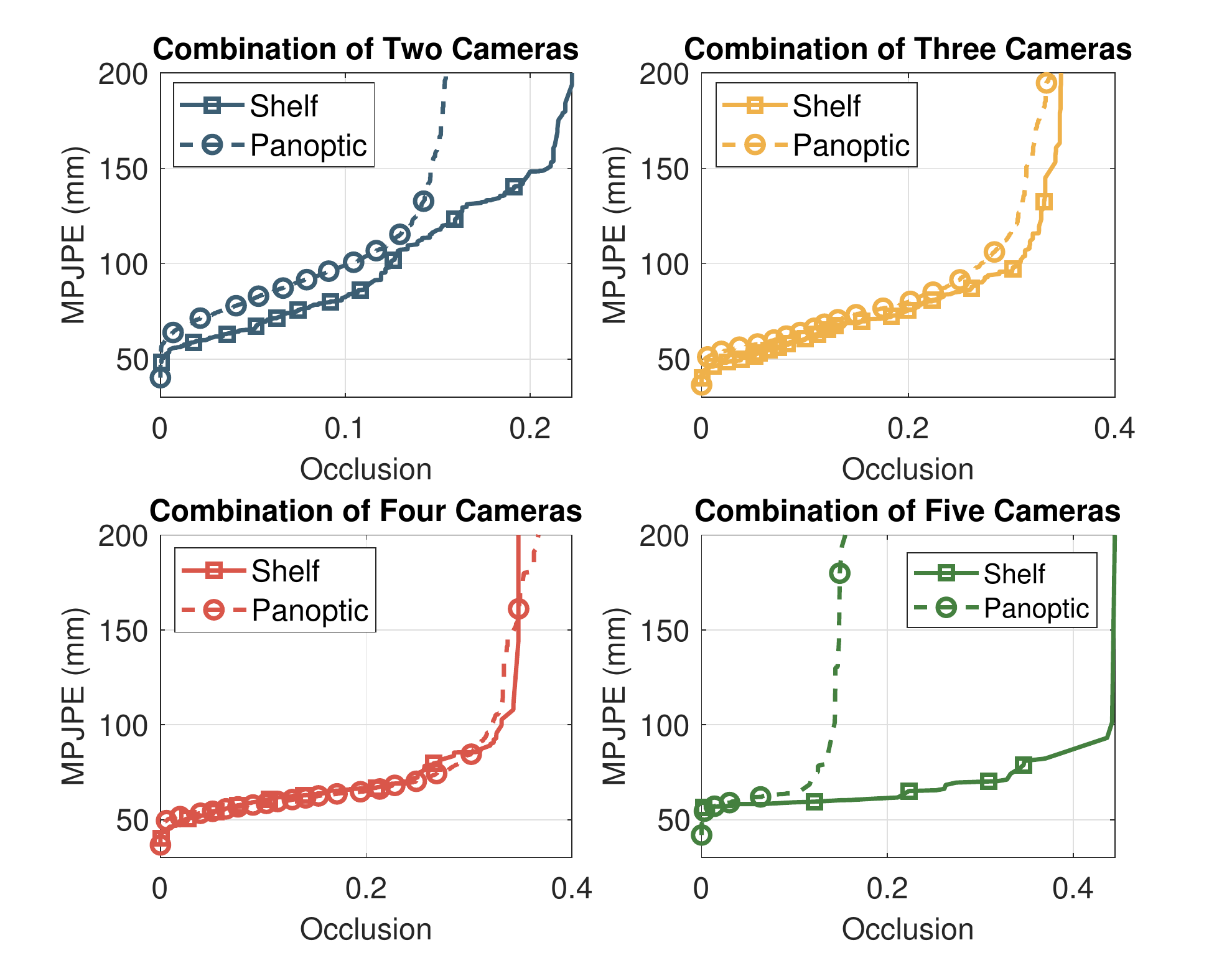}
	\vspace{-25 pt}
	\caption{Relation of the occlusion (i.e., IoU) and the 3D pose estimation accuracy (i.e., MPJPE).}
	\label{fig:occlusionAccuracy}
\end{figure}

Let $s^t_n \in \{0, 1\}$ be the binary selection variable for camera $n$ at time slot $t$ where $s^t_n = 1$ means that camera $n$ is selected and $s^t_n = 0$ otherwise. The goal of the camera scheduler is to select a subset of $C$ cameras so that the total occlusion of the selected cameras is minimized. Note that ideally the objective function should be to minimize the 3D pose estimation error. However, it is extremely difficult, if not impossible to construct a 3D pose estimation error function depending on the selected cameras and even their estimated 2D poses. As such, we minimize the total occlusion of the selected cameras instead, which is much easier to calculate given the 2D bounding boxes in each camera views. In our experiments on the Shelf dataset \cite{belagiannis20143d} and the Panoptic dataset \cite{joo2015panoptic}, we found that the 3D pose estimation error is monotonically related to the total occlusion, thereby justifying the choice of our objective function. Specifically, Figure \ref{fig:occlusionAccuracy} illustrates the MPJPE as a function of the total occlusion for different numbers of cameras in the selected subset. In all cases, the MPJPE monotonically increases with the IoU.

Formally, the camera scheduler aims to solve the following long-term optimization problem for a number of $T$ time slots:
\begin{equation}
    \setlength{\abovedisplayskip}{3pt}
    \setlength{\belowdisplayskip}{3pt}
	\label{eq:objectiveFunc}
	\begin{aligned}
		\min_{s^1, ..., s^T}~~ &\frac{1}{T} \sum_{t=1}^{T} \sum_{n=1}^{N} \text{IoU}_n^t \cdot s_n^t\\
		s.t.~~& \frac{1}{T} \sum_{t=1}^{T} s_n^t \le E_n, n=1, \dots, N \\
		&\sum_{n=1}^{N} s_n^t = C, \forall t
	\end{aligned}
\end{equation}
where the first constraint is a long-term energy constraint determined by each camera's (normalized) battery energy capacity $E_n$ and the second constraint specifies the number of cameras to select in each time slot. Note that $E_n$ must satisfy $E_n \ge C/N$. There are two main challenges that impede the derivation of the optimal solution of the scheduling problem. 

Firstly, although IoU is a meaningful metric to measure occlusion, it can only be calculated when the 2D bounding boxes in a camera are available. However, this is a chicken-egg problem because the 2D bounding boxes cannot be obtained without having the camera perform the 2D pose estimation on the frame in the first place. 

Secondly, optimally solving the above optimization problem requires the information of all future scenes, which is impossible to know in advance. Moreover, the long-term energy constraint couples the camera selection decision across time: consuming more energy in the current time slot will reduce the available energy for future use. Therefore, a greedy algorithm that myopically minimizes the occlusion in the current time slot may create processing hotspots on some cameras, thereby quickly depleting their battery energy and reducing the battery lifetime of the overall system. 

Next, we describe our designs to cope with these challenges. 

\subsubsection{\textbf{How to calculate IoU without performing 2D pose estimation?}}
To calculate the IoUs before cameras are selected to perform 2D pose estimation on their captured images, our idea is to let the edge server \textit{predict} the 3D poses based on the past 3D pose estimation results, and then project the predicted 3D poses onto each camera's view to obtain the 2D poses. Using the projected 2D poses (and their corresponding bounding boxes), the camera scheduler then calculates the IoU for each camera. Because the IoU calculation is executed completely by the edge server and does not require any computation or message exchange at the camera side, cameras incur no extra power consumption while the transmission delay is eliminated.

\textbf{Fast 3D Pose Prediction}. We design an attention-based LSTM as the 3D pose predictor because of its powerful capabilities of learning rich spatial and temporal features from a series of historical 3D poses. In addition, it incurs a very low prediction delay. The experiments on our testbed report a 5.6ms prediction delay on average. In time slot $t$, the camera scheduler aims to predict the real-world coordinates of all 3D poses in the scene in a future time slot $t+\tau$, given historical estimated 3D poses of the past $M$ time slots. The time advance $\tau$ accounts for the delay incurred during prediction and transferring the control signals to the cameras, and ensures that the cameras are informed of the selection decision before the scene of interest comes up. Moreover, $\tau$ needs not be a fixed value but is adaptive to the deployment scenario and the dynamic wireless channel conditions. Thus, the predictor may predict the 3D poses several time slots beyond the immediate next time slot. 

\begin{figure}[tt]
	\centering
	\includegraphics[width=\linewidth]{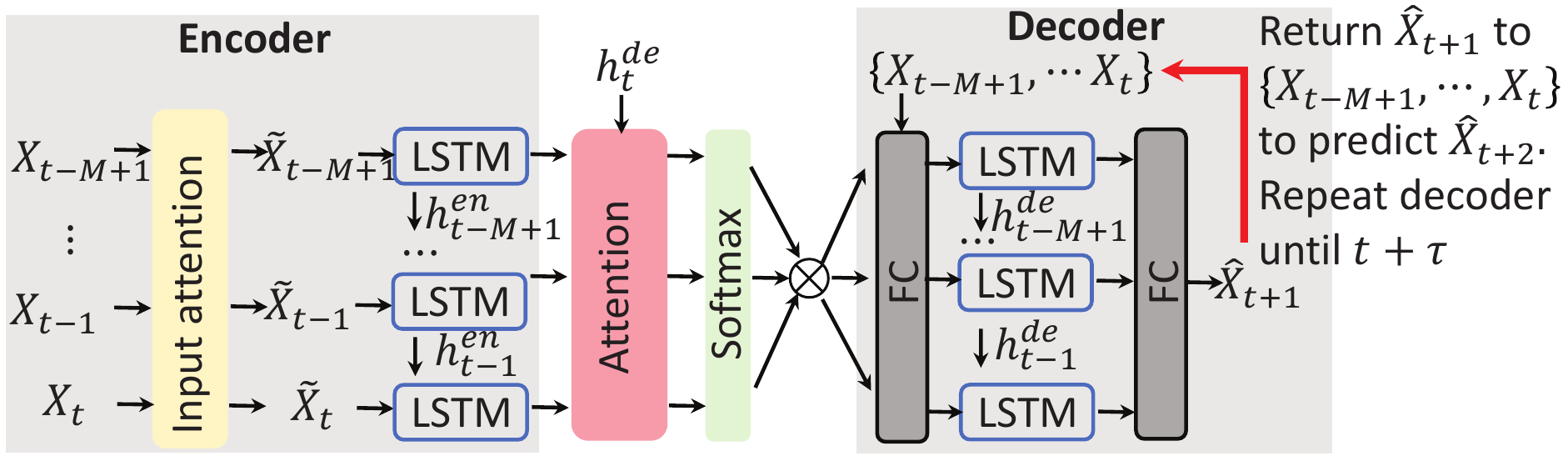}
    \vspace{-20 pt}
	\caption{The adopted attention-based LSTM network. We modify this network by adding a feedback loop to enable the prediction for the scene at slot $t + \tau$.}
	\label{fig:predictionModel}
\end{figure}

Mathematically, we consider a loss minimization problem for training the attention-based LSTM network, denoted by $f(\cdot; \theta)$ parameterized by $\theta$, as follows:
\begin{equation}
    \setlength{\abovedisplayskip}{3pt}
    \setlength{\belowdisplayskip}{3pt}
	\begin{aligned}
		& \min_\theta \mathcal{L}(\hat{\mathcal{X}}, \mathcal{X}) = \min \sum_{l=1}^{\tau} (\hat{X}^{t+l} - X^{t+l})^2 \\
		& s.t. \hat{\mathcal{X}} = f(\{X^{t-M+1}, X^{t-M+2}, \dots, X^t\}; \theta)
	\end{aligned}	
\end{equation}
where $\hat{\mathcal{X}} = \{\hat{X}^{t+1}, \dots, \hat{X}^{t+\tau}\}$ is the predicted 3D poses, and $\mathcal{X} = \{X^{t+1}, \allowbreak \dots, X^{t+\tau}\}$ is the ground truth 3D poses.

Recently, attention-based LSTM models \cite{zhang2021elf, yuan2020using, zeng2022satellite} have shown their effectiveness in predicting time series data. As shown in Figure \ref{fig:predictionModel}, our attention-based LSTM network consists of three modules: an encoder, an attention module and a decoder. The encoder takes the spatial and temporal information (i.e., the 3D poses' coordinates in the past $M$ time slots) of each human pose $\{X^{t-M+1}, ..., X^{t}\}$ as input, and encodes them into the feature map by passing them through the input attention module and the LSTM module. Subsequently, the attention module, which is a fully-connected layer, is adopted to select the most relevant encoded features and generate the context vector. Finally, the decoder processes the context vector associated with the historical 3D poses through a fully-connected layer, a LSTM model and a fully-connected regressor to output the final 3D pose prediction. Traditional dual-stage attention-based LSTM models usually only make the prediction for the immediate next time slot $t+1$. In our design, in order to make the prediction for a future time slot $t + \tau$, we feed the prediction $\hat{X}$ back to the decoder $\tau$ times to yield the predicted 3D poses $\hat{X}^{t+\tau}$ in time slot $t + \tau$. 

\textbf{Projection}. Next, the predicted 3D poses $\hat{X}^{t+\tau}$ are projected onto each camera $n$'s view to obtain the predicted 2D poses $\hat{x}_n^{t+ \tau}$ using camera $n$'s projection matrix $P_n$ as follows,
\begin{equation} \label{eq:reprojection}
	q [\hat{x}_n^{t+ \tau}, 1]^\top = P_n[
	\hat{X}^{t+ \tau},
	1]^\top
\end{equation}
where $q$ is the scaling constant between the pixel coordination system and the world coordination system. 

\textbf{Occlusion Calculation}. With the projected 2D poses, the camera scheduler generates the bounding boxes of each 2D pose in each camera's view. However, because of the discrepancy between the 2D pose and the actual human body in an image, we slightly expand (by 10\% in our implementation) the bounding box of the 2D pose to obtain the bounding box of the human object before calculating the occlusion in each camera view. Using the expanded bounding boxes, the IoU in each camera view is calculated by Equation \eqref{eq:iouCalculation}. 


\subsubsection{\textbf{How to make the camera selection decisions without future information?}} \label{sec:makeCameraDecision}
We leverage the Lyapunov optimization technique \cite{neely2010stochastic} to make camera selection decisions without knowing far future information to balance long-term 3D estimation performance and energy consumption. Specifically, \EP~ converts the long-term optimization problem \eqref{eq:objectiveFunc} into a sequence of per-slot optimization problems that can be easily solved using the predicted occlusion information and the current energy state. To handle the long-term energy constraint that couples the camera selection decisions across time slots, we construct a (virtual) energy deficit queue for each camera to guide the camera selection decision to follow the long-term energy constraint. Let $q^t_n$ denote the energy deficit queue length of camera $n$ in time slot $t$ with $q^0_n = 0$ as follows,
\begin{equation}
    \setlength{\abovedisplayskip}{3pt}
    \setlength{\belowdisplayskip}{3pt}
	q_n^{t+1} = \max \{q_n^t - E_n, 0\}  + s_n^t
\end{equation}
Thus $q^t_n$ indicates the deviation of the current power consumption from the long-term energy constraint. Following the drift-plus-penalty framework in Lyapunov optimization \cite{neely2010stochastic}, in time slot $t$, the camera scheduler makes the selection decision for the scene in time slot $t + \tau$ by solving the following optimization problem
\begin{equation}
    \setlength{\abovedisplayskip}{3pt}
    \setlength{\belowdisplayskip}{3pt}
	\label{eq:lyapunovDrift}
	\begin{aligned}
	 \min_{\textbf{s}^t}~~ &\sum_{n=1}^N \left( V \cdot \text{IoU}_n^{t+\tau} s_n^{t+\tau} + q_n^{t+\tau} \cdot s_n^{t+\tau} \right) \\
	 s.t.~~ &\sum_{n=1}^{N} s_n^{t+\tau} = C
	\end{aligned}
\end{equation}
Apparently, the optimal solution can be easily obtained by first calculating $V\cdot \text{IoU}^{t+\tau}_n + q^{t + \tau}_n$ for each camera $n$, and then selecting $C$ cameras with the smallest values. The first term in the objective function \eqref{eq:lyapunovDrift} is to minimize the occlusion of the selected cameras and the second term is added to satisfy the long-term energy constraint. The positive control parameter $V$ is used to adjust the trade-off between these two purposes. In particular, by considering the additional term $q^{t+\tau}_n$, the camera scheduler takes into account the (time-varying) energy deficit of the camera: when $q^{t + \tau}$ is larger, minimizing the energy deficit is more critical and hence cameras with a larger $q^{t + \tau}$ will be less likely to be selected. Thus, the scheduler works by following the philosophy of ``if violate the energy budget, then use less energy'', and hence the long-term energy constraint can be satisfied in the long-run without foreseeing all future information. 
We summarize the camera scheduling algorithm in Algorithm \ref{alg:alorithm1}. 

\begin{algorithm}
	\caption{Algorithm 1}
	\begin{algorithmic}[1]
		\State Given control parameter $V$, energy deficit queues $\mathbf{q}^0 = 0$.
		\For {$t=0$ to $t = M-1$}
		    \State Estimate the 3D poses using all cameras.
		\EndFor
		\For {$t=M$ to $t = T$}
			\State Estimate the 3D poses based on the selected cameras.
			\State Predict the 3D poses in time slot $t+\tau$;
			\State Solve problem \eqref{eq:lyapunovDrift} to get the camera selection decision in time slot $t+\tau$.
			\State Update the deficit $\mathbf{q}^{t+\tau}$ for all cameras:
			\State $q_n^{t+\tau+1} = \max \{q_n^{t+\tau} - E_n\} + s_n^{t+\tau} $
			\State Send the camera selection decision to each camera.
		\EndFor
	\end{algorithmic}
	\label{alg:alorithm1}
\end{algorithm}

\subsection{Discussion on the Choice of Method}
\begin{figure*}[tt] 
	\centering
	\subfigure[$E^3$Pose]{
		\label{fig:predictionTimeline}
		\includegraphics[width=0.3\linewidth]{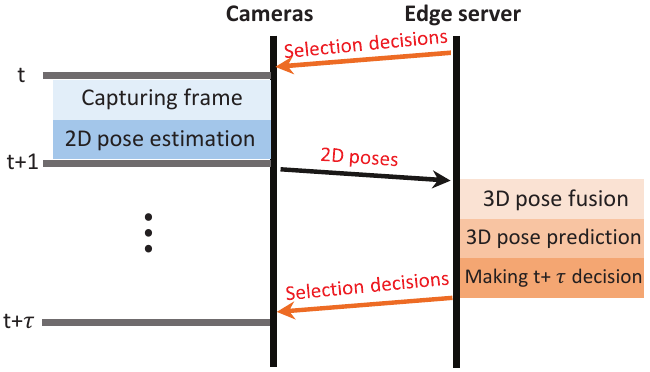}}
	\hspace{0.02\linewidth}
	\subfigure[Central control with occlusion information from 2D bounding boxes prediction]{
		\label{fig:centralControlBB}
		\includegraphics[width=0.3\linewidth]{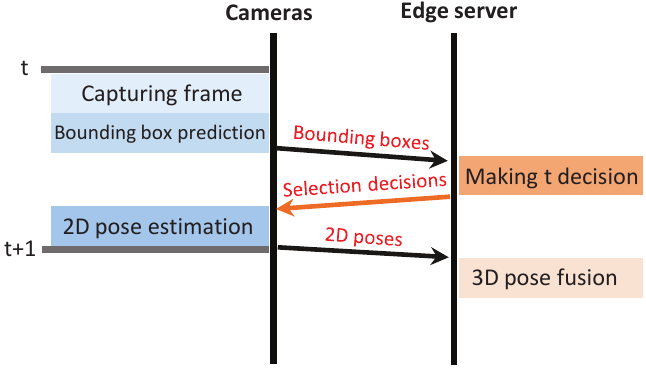}}
	\hspace{0.02\linewidth}
	\subfigure[Distributed control with occlusion information from 2D bounding boxes prediction]{
		\label{fig:distributedControlBB}
		\includegraphics[width=0.3\linewidth]{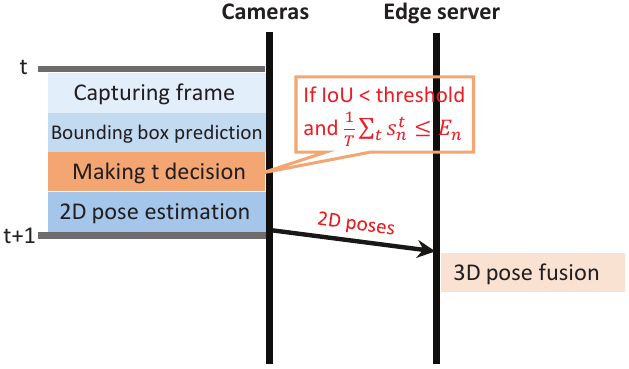}}
    \vspace{-15 pt}
	\caption{Comparison of different system frameworks. Due to the use of prediction-based camera selection, \EP~ is also able to achieve a higher frame rate.}
	\label{fig:timelineComparison}
	\vspace{-10 pt}
\end{figure*}

The workflow of \EP~ is shown in Figure \ref{fig:predictionTimeline}. A key innovation of our proposed \EP~ system is to predict future 3D poses and then project the prediction results onto each individual camera view to obtain the 2D bounding boxes to enable the IoU calculation. This is because the IoU must be calculated before 2D poses in the scene of interest are estimated. In what follows, we discuss the rationale of our proposed prediction method. 

\textbf{Why not directly perform 2D bounding box prediction?} Since the purpose of 3D pose prediction is to calculate the IoU in the 2D camera view, an alternative approach is to directly predict the future 2D bounding boxes using the current 2D bounding boxes in each camera view without performing 3D pose prediction, using, e.g., 2D motion vector techniques \cite{liu2019edge, zhang2020mobipose}. We argue that this approach is not suitable for real-time multi-human 3D pose estimation in a multi-camera system. Consider the two possible implementations of this alternative approach. In the first implementation as shown in Figure \ref{fig:centralControlBB}, each camera performs the bounding box prediction, and sends the prediction results to the edge server which then makes the camera selection decision. In this implementation, not only the camera is assigned with the extra work of bounding box prediction (which entails extra energy consumption and delay), but also more wireless network bandwidth is consumed because of the extra message exchange. Because the cameras need to wait for the selection decisions of the current scene to come back, the frame rate will also be limited by the roundtrip transmission time and the edge server processing speed. In the second implementation as shown in Figure \ref{fig:distributedControlBB}, each camera performs the bounding box prediction, calculates the IoU, and decides whether to perform 2D pose estimation based on the calculated IoU by itself (e.g., using a threshold method). In this implementation, camera selection is a fully distributed decision and hence fewer control messages are exchanged. However, because of the lack of central coordination, there is no guarantee on the number of cameras that perform 2D pose estimation or an efficient trade-off between the 3D pose estimation accuracy and the system energy efficiency. For example, it is entirely possible that not a single camera decides to perform 2D pose estimation due to the autonomous decision making, thereby resulting in a 3D pose estimation failure. 

\textbf{Why not use 3D motion vector to predict 3D poses?}
To predict the 3D poses, one may alternatively use a 3D motion vector-based predictor instead of our proposed attention-based LSTM. However, our experiments show that 3D motion vectors can achieve a good prediction result for the immediate next time slot, but fails to perform well for time slots further into the future. This is because 
3D motion vectors represent a transient motion trajectory, and there will be a large error if this transient motion trajectory is superimposed to obtain the future 3D poses after a relatively longer period of time. Hence, \EP~ uses a light-weight attention-based LSTM to predict the future 3D human poses, which can learn the long-range motion dependency from historical 3D human poses and use it to get a better prediction result. To further compare the 3D pose prediction performance by using 3D motion vectors and the attention-based LSTM, we incorporate 3D motion vectors into \EP~ and use it as a baseline in Section \ref{sec:performanceOfOcclusionPrediction}.

\textbf{Why not use the results of 3D pose prediction as the final 3D human poses?} \EP~ uses 3D pose prediction to calculate the occlusions in the 2D camera views, thereby guiding the camera selection. Thus, 3D pose prediction only generates coarse 3D poses, whose accuracy is not comparable with the 3D poses estimated using the realized images of the scene.

\section{System implementation}

We implement a prototype of \EP~ in Python for easy integration with deep learning modules. Our implementation adopts a multi-thread parallel method with four camera threads and four edge server threads which we introduce in more detail below.

\textbf{Camera Threads}. The Video Capture thread uses OpenCV \cite{bradski2008learning} to capture the live video stream from the on-device camera and puts the frames into the frame queue. In the 2D Pose Estimation thread, we adopt the CNN architecture in \cite{cao2017realtime} with significantly more lightweight ResNet18 \cite{he2016deep} as the backbone feature extractor to obtain the 2D poses. The 2D pose estimation model is first trained offline on the COCO dataset \cite{lin2014microsoft} using PyTorch. After training, the model is converted to a TensorRT-compatible model \cite{vanholder2016efficient} since TensorRT enables on-device machine learning inference with low latency and small binary size. The 2D Pose Estimation thread gets frames from the global frame queue and runs 2D pose estimation if the camera is selected for the current frame. The detected 2D poses are then put into the 2D pose queue. The network communication is implemented using the Socket library \cite{van1995python} in Python in two separate threads: Client Sending thread and Client Receiving thread. The Client Sending thread sends the 2D pose results to the edge server via wireless and the Client Receiving thread receives the camera selection decision from the edge server and updates the camera mode switch. 

\textbf{Edge Server Threads}. In the 3D Pose Fusion thread, we use an efficient iterative greedy matching algorithm proposed in \cite{tanke2019iterative} to associate the 2D poses across multiple camera views, and use the triangulation method in computer vision to build the human 3D poses. The 3D pose results are stored in the 3D pose queue. In the Camera Selection thread, we modify the attention-based LSTM network \cite{qin2017dual} for 3D pose predictions and solve the optimization problem \eqref{eq:lyapunovDrift} to obtain the camera selection decision. The network communication at the edge server is also separated into two threads: Server Sending thread and Server Receiving thread. The Server Sending thread gets the camera selection result from the decision queue and sends it to each camera. The Server Receiving thread receives the 2D pose results from the cameras and store them into the multi-view 2D pose queue. 


\section{Evaluation}
\subsection{Experiment Setup}
\begin{figure}[tt]
	\centering
	\includegraphics[width=\linewidth]{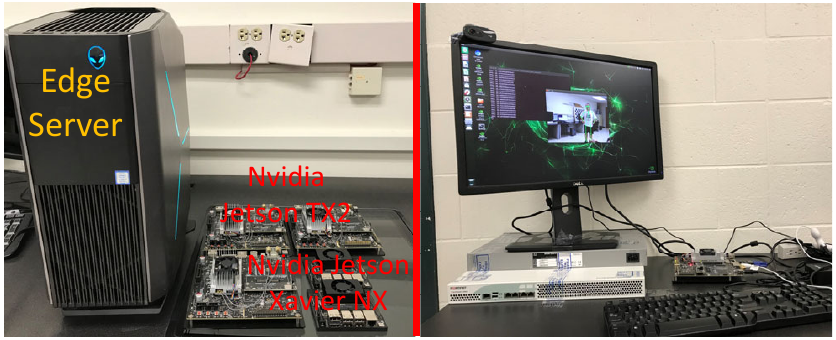}
    \vspace{-20 pt}
	\caption{Evaluation testbed. Left: the edge server and the camera devices. Right: The setup of a camera and the runtime monitor.}
	\label{fig:systemSnapshot}
\end{figure}

\textbf{Hardware Testbed}. We build a prototype of \EP~ on a hardware testbed consisting of five cameras and one edge server. We use two NVIDIA Jetson Xavier NX (referred to as Xavier from now on) devices and three NVIDIA Jetson TX2 (referred to as TX2 from now on) devices as the smart cameras. Each camera is equipped with a Logitech C270 HD Webcam. To remove the impact of DVFS (Dynamic Voltage and Frequency Scaling) \cite{le2010dynamic} and allow repeatable experiments, the CPU and the GPU are set at the highest frequencies on all the camera devices. A Dell desktop computer is employed as the edge server, which is equipped with an Intel Core i7-8700K CPU at 3.70GHz$\times12$, two NVIDIA GeForce GTX 1080 Ti GPUs, and 11 GB memory. The smart cameras and edge server are wirelessly connected by WiFi. Figure \ref{fig:systemSnapshot} shows a picture of the server and a camera device on our testbed.

\textbf{Dataset}. Two public multi-camera datasets are used to evaluate the performance of $E^3$Pose. \textbf{Shelf} \cite{belagiannis20143d}. The Shelf dataset contains 3200 video frames captured by five cameras in an indoor environment with four persons interacting with each other. We follow the previous works \cite{tanke2019iterative,qiu2019cross, remelli2020lightweight, chen2020cross, dong2019fast} to evaluate the accuracy of 3D pose estimation. \textbf{Panoptic} \cite{joo2015panoptic}. The Panoptic dataset is captured in a closed studio with 480 VGA cameras and 31 HD cameras. The hundreds of cameras are distributed over the surface of a geodesic sphere with about 5 meters of width and 4 meters of height. The studio is designed to simulate and capture social activities of multiple people. We use the same set of training and testing sequences captured by the same set of five HD cameras (3, 6, 12, 13, 23) as in \cite{tu2020voxelpose, lin2021multi} for evaluation.

\textbf{Live Videos from Camera}
We also develop a test case that runs \EP~ on our hardware testbed over the live videos captured by the Logitech C270 HD Webcams. The number of people in the live videos ranges from two to three.  For offline training the attention-based LSTM network in this case, five video clips are captured from different view and recorded. At runtime, we use live videos to evaluate the power consumption of the system. 

\textbf{Training the 3D Pose Prediction Model.}
The attention-based LSTM model for 3D pose prediction is trained offline. Although the Shelf dataset has 3200 frames, only 280 frames have the 3D human pose ground truth information. As a result, we do not have enough ground truth data as the training data. Similar issues also exist for the Panoptic dataset. To overcome this issue, we create an expanded training dataset using the estimated 3D poses as the ground truth. Specifically, for each scene in each time slot, 3D human poses are estimated (in an offline fashion) by all possible camera combinations. Each training input thus is the 3D poses of all human objects in the scene for $M$ consecutive time slots, where each human object's 3D pose in a time slot is selected randomly from the expanded training dataset in the corresponding time slot. For our live videos, due to the lack of ground truth, we pick the subset of cameras with small occlusion and use their estimated 3D poses as the ground truth. 
Note that the attention-based LSTM can also be trained/updated online as more sequences of 3D pose estimates are generated by the system over time. 

\textbf{Baselines.}
\EP~ is compared with the following baselines:
\textbf{Select-All (SA)}: All cameras perform 2D pose estimation for each video frame and send the results to the edge server for 3D pose fusion. No camera scheduling is performed. 
\textbf{Random Selection (RS)}: The edge server  simply randomly chooses 3 cameras to perform the 2D pose estimation and synthesizes the results to estimate 3D poses. 
\textbf{Independent Decision (ID)}: This is the baseline shown in Figure \ref{fig:distributedControlBB}. Each camera uses the motion vector method in \cite{liu2019edge} to predict the 2D bounding boxes and calculates the IoU based on the predicted 2D bounding boxes. Then the cameras decide independently by themselves whether to perform 2D estimation in the future scene of interest. 
\textbf{\EP~ with 3D motion vector prediction (\EP-MV)}: Instead of using the attention-based LSTM to predict 3D poses, this baseline uses a 3D motion vector method to predict the 3D poses and incorporate them into \EP.

\textbf{Evaluation Metrics}
We evaluate \EP~ in terms of the power consumption and the 3D pose estimation accuracy.
\textbf{Power consumption}: We utilize the INA3221 power monitor \cite{nvidiaPower} embedded in TX2 and Xavier to read the power consumption during runtime. We use the average power consumption for a scene as our evaluation metric. The power consumption directly translates to system battery lifetime. 
\textbf{3D Pose Estimation Accuracy}: 
For the Shelf dataset, we use the Percentage of Correctly estimated Parts (PCP) as a metric to evaluate the accuracy of the estimated 3D poses to enable a direct comparison with existing works \cite{tanke2019iterative,qiu2019cross, remelli2020lightweight, chen2020cross, dong2019fast}. For the Panoptic dataset, since existing works do not have a common evaluation protocol, we extend the Average Precision $(AP_K$) metric \cite{pishchulin2016deepcut} to the multi-person 3D pose estimation problem, which is defined as the percentage of estimated 3D poses whose MPJPE is smaller than $K$ millimeters. 
    

\subsection{Performance of 3D Pose Estimation}

\textbf{Estimation Accuracy.} We first compare \EP~ with existing methods on both Shelf dataset and Panoptic dataset in Table \ref{table:accuracyCompareOthers} and Table \ref{table:accuracyCompareBaselinesCMU}. We point out that these existing methods use \textit{all} 5 cameras to perform 3D pose estimation for each scene while \EP~only adaptively selects 3 cameras. Although this is not a fair comparison for \EP, we can see that \EP~achieves a comparable estimation accuracy with the state-of-the-art solutions. Among these methods, \cite{belagiannis20143d,belagiannis2014multiple,dong2019fast,chen2020cross,lin2021multi,tu2020voxelpose} study 3D pose estimation from a pure computer vision view without considering system design. 
The work in \cite{simon2021} proposes an edge-assisted multi-camera system similar to ours but utilizes all cameras to perform 3D pose estimation. In fact, the baseline SA can be considered as a simplified version of \cite{simon2021}. The difference is that \cite{simon2021} in addition feeds the 3D pose estimation result from the edge server back to the individual cameras to improve the 2D pose estimation performance in the next time slot. However, this closed-loop design prohibits pipeline processing and can result in a reduced frame rate when the server processing speed is low or the wireless condition is bad. \EP~as well as the considered SA baseline use an open-loop design where the cameras do not need to wait for the 3D pose results from the edge server to start the processing of the next frame. 

\begin{table}[h]
	\caption{Performance comparison on the Shelf dataset with existing methods and the baselines: Percentage of Correct Parts (PCP) (\%).}
  \vspace{-10 pt}
	\begin{center}
		\begin{tabular}{c|cccc}
			\toprule[1.5pt]
		    & \multicolumn{4}{c}{\textbf{PCP (\%)}}\\
			\cline{2-5}
		    & Actor1 & Actor2 & Actor3 & Avg\\
			\hline
		    CVPR2014 \cite{belagiannis20143d}& 66.1 & 65.0 & 83.2 & 71.4  \\
			ECCV2014 \cite{belagiannis2014multiple}  & 75.0 & 67.0 & 86.0 & 76.0 \\
            CVPR2019 \cite{dong2019fast}  & 98.8 & 94.1 & 97.8 & 96.9\\
            CVPR2020 \cite{chen2020cross} & 99.6 & 93.2 & 97.4 & 96.7\\
            CVPR2021 \cite{lin2021multi} & 99.3 & \textbf{96.5} & 98.0 & \textbf{97.9}\\
            RSS2021 \cite{simon2021} & 99.3 & 95.7 & 97.3 & 97.4 \\
            \hline
            SA & 98.7 & 86.7 & 97.7 & 94.4\\
		    RS & 95.8 & 83.2 & 95.5 & 91.5 \\
		    ID & 89.1 & 78.6 & 85.6 & 84.4 \\
		    \EP-MV & 97.2 & 84.5 & 96.9 & 92.9\\
            \EP & \textbf{99.9} & 91.9 & \textbf{99.4} & 97.1 \\
			\toprule[1.5pt]
		\end{tabular}
		\label{table:accuracyCompareOthers}
	\end{center}
	\vspace{-10 pt}
\end{table}

In Table \ref{table:accuracyCompareOthers} and Table \ref{table:accuracyCompareBaselinesCMU}, we also perform comparison with the considered baselines on the Shelf dataset and the Panoptic dataset, respectively. The results show that \EP~ outperforms the baselines by 2.8\% to 13.1\% in terms of PCP on Shelf and 19.7\% to 41.14\% in terms of MPJPE on Panoptic. We explain the results in more details next. \textbf{SA}: Although the edge server receives all the 2D poses from the smart cameras, errors in the 2D estimation results with severe occlusions can bring down the accuracy of 3D pose estimation. This reconfirms our findings in Section \ref{sec:accuracy_3dfusion}. Also note that SA achieves a slightly lower PCP accuracy than \cite{simon2021} as expected since the feedback mechanism is not used. \textbf{RS}: Since cameras are randomly selected to perform 2D pose estimation, cameras with low view qualities (i.e., large occlusion) can be selected and their poor 2D pose estimation results can be included in the 3D pose synthesis. \textbf{ID}: Because cameras make decisions by themselves without any central coordination, it is possible that too few or even zero cameras decide to perform 2D pose estimation in some time slots, thereby reducing the 3D estimation accuracy. \textbf{\EP-MV}: Instead of using an attention-based LSTM, \EP-MV uses the 3D motion vectors of human joints to predict the future 3D poses. However, the transient motion trajectory is not effective in predicting the 3D poses after a relatively long-period of time (e.g., several time slots beyond the current scene) since the movement of human joints can be potentially variable. This leads to poor occlusion prediction results and hence reduced 3D pose estimation accuracy. We will show more 3D pose estimation performance results shortly. 

\begin{table}[tt] 
	\caption{Performance comparison on the Panoptic dataset with the baselines: Average Precision ($AP_K$) (\%).}
  \vspace{-10 pt}
	\begin{center}
		\begin{tabular}{c|cccc}
			\toprule[1.5pt]
		    & $AP_{25}$ & $AP_{50}$ & $AP_{105}$ & $AP_{150}$\\
			\hline
            ECCV2020 \cite{tu2020voxelpose} & 83.5 & 98.3 &99.7 & 99.9 \\
            CVPR2021 \cite{lin2021multi} & \textbf{92.1} & \textbf{98.9} & 99.8 & 99.8 \\
            \hline
            SA & 86.3 & 93.3 & 95.5 & 96.8 \\
		    RS & 66.3 & 83.7 & 91.3 & 91.3 \\
		    ID & 40.6 & 69.3 & 86.2 & 90.3  \\
		    \EP-MV & 76.7 & 94.0 & 98.0 & 98.4 \\
            \EP & 90.5 & 98.6 & \textbf{99.8} & \textbf{100.0}  \\
			\toprule[1.5pt]
		\end{tabular}
		\label{table:accuracyCompareBaselinesCMU}
	\end{center}
	\vspace{-10 pt}
\end{table}

\begin{figure}[tt]
	\centering
	\begin{minipage}[t]{0.48\linewidth}
		\includegraphics[width=\textwidth]{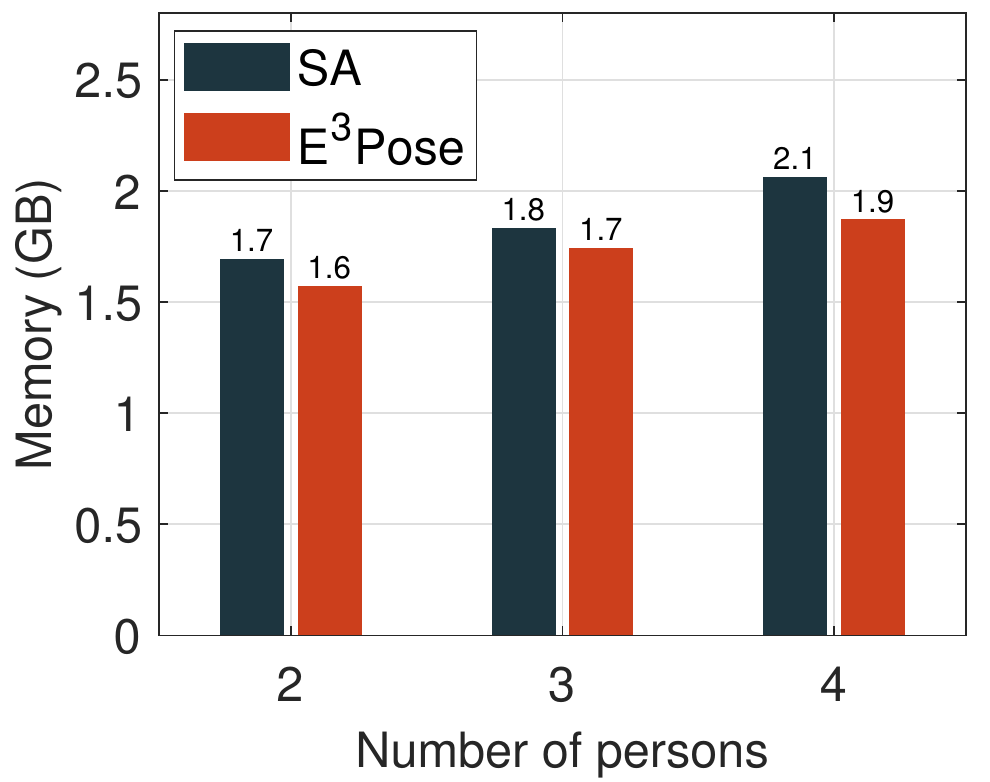}
		\vspace{-20 pt}
		\caption{Memory usage under different number of persons in the scene.}
		\label{fig:memoryUsage}
	\end{minipage}
	\hspace{0.01\linewidth}
	\begin{minipage}[t]{0.48\linewidth}
		\includegraphics[width=\textwidth]{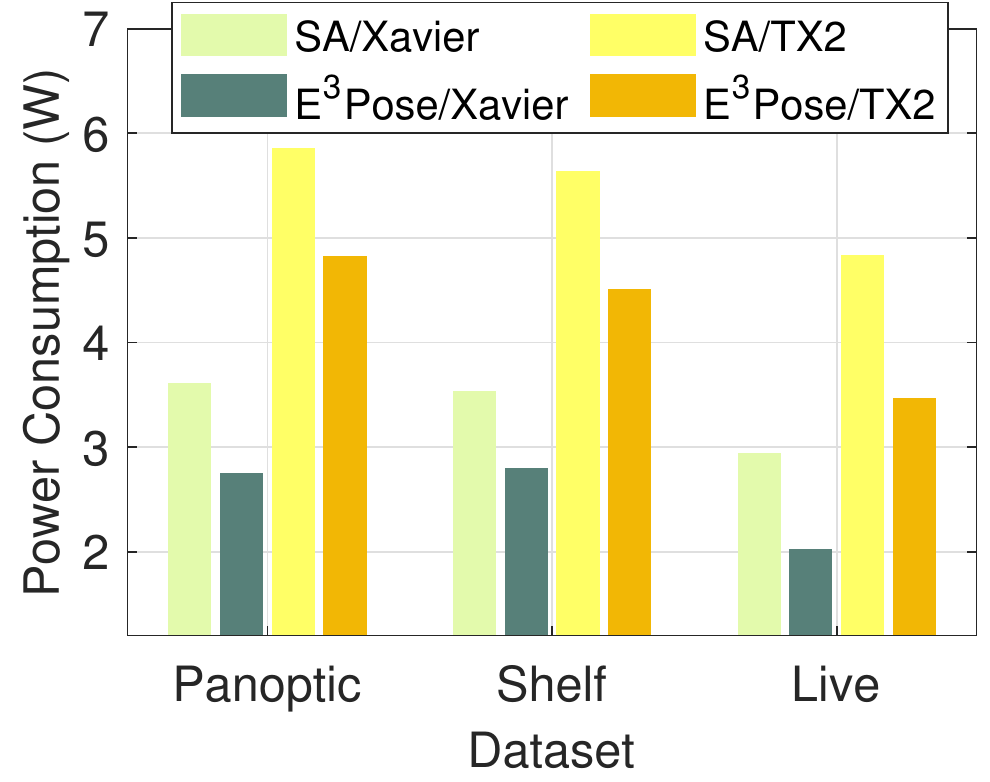}
		\vspace{-20 pt}
		\caption{Power consumption of Xavier and TX2 on three different datasets.}
		\label{fig:powerXavierAndTX2}
	\end{minipage}
\end{figure}

\begin{figure}[tt]
	\centering  
	\subfigure[NVIDIA Jetson Xavier NX]{
		\label{fig:powerDetailOnXavier}
		\includegraphics[width=0.48\linewidth]{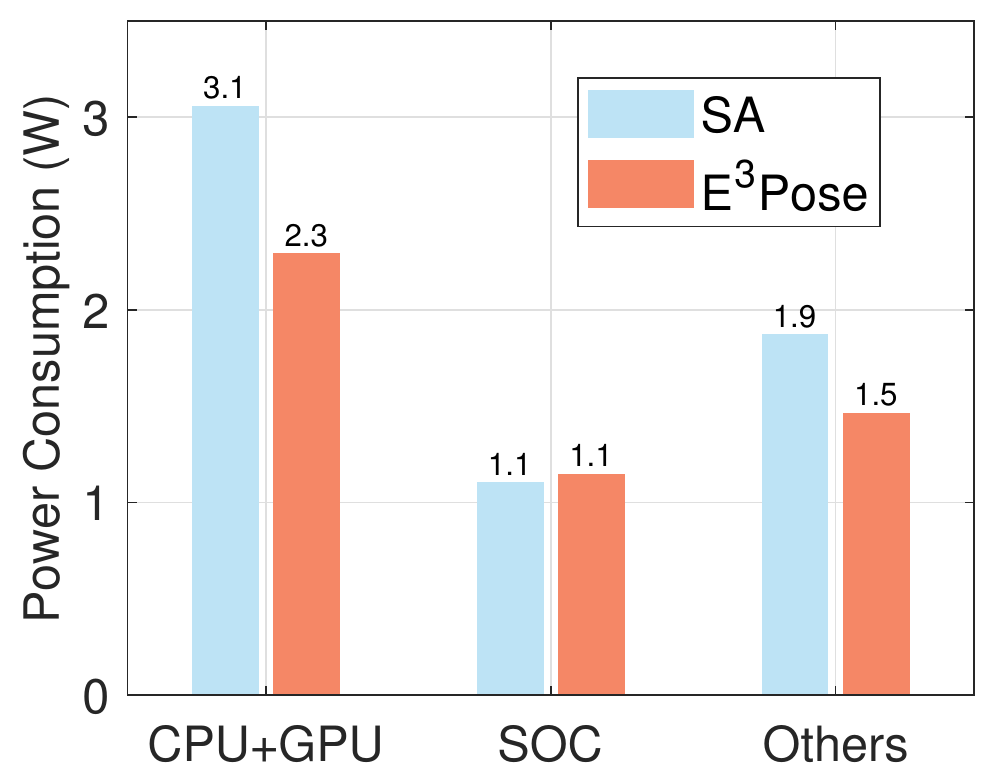}}
	\subfigure[NVIDIA Jetson TX2]{
		\label{fig:powerDetailOnTX2}
		\includegraphics[width=0.48\linewidth]{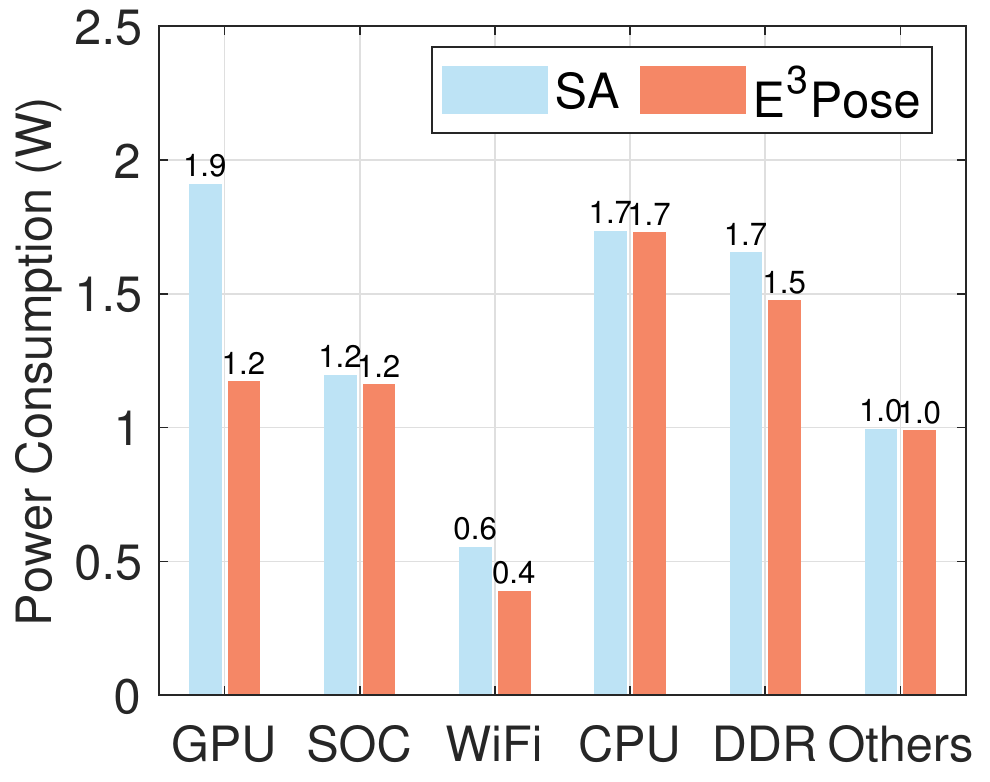}}
	\vspace{-15 pt}
	\caption{Power consumption breakdown on Xavier and TX2.}
	\label{fig:powerDetail}
	\vspace{-10 pt}
\end{figure}


\textbf{Memory Usage.}
We also measure the memory usage of SA and \EP~ using the monitor tools in the NVIDIA devices. Figure \ref{fig:memoryUsage} shows that, as expected, \EP~ incurs a similar memory usage with SA (\EP~ being just slightly lower) because even cameras in the PSM still have the DNN stored in the memory for quick mode switching. Figure \ref{fig:memoryUsage} further shows that the memory overhead slightly increases with the number of persons in the scene, since more instances are created for 2D pose estimation. 
\begin{figure}[tt]
	\centering
	\includegraphics[width=0.9\linewidth]{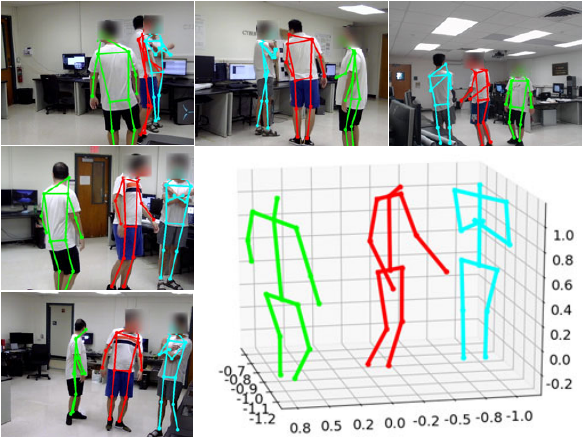}
    \vspace{-10 pt}
	\caption{Example of \EP~ on live videos from cameras. }
	\label{fig:liveVideoResult}
\end{figure}

\textbf{Power Consumption.}
We analyze the power consumption of Xavier and TX2 devices separately due to their different hardware and system architectures. In Figure \ref{fig:powerXavierAndTX2}, we show the power consumption of Xavier and TX2 on two datasets and the live video, respectively. The power consumption here includes the power consumption by GPU, CPU, WiFi, DDR, but does not include SOC and others, whose changes are negligible between SA and \EP. As can be seen, \EP~ can save 20.66\% to 31.21\% power on Xavier and 17.72\% to 28.67\% power on TX2.

Figure \ref{fig:powerDetail} shows the breakdown power consumption of the different system components of Xavier and TX2 on the Shelf dataset. As Figure \ref{fig:powerDetailOnXavier} shows, \EP~ utilizes less energy in CPU+GPU and others (DDR and WiFi) than SA for Xavier.  Figure \ref{fig:powerDetailOnTX2} shows a more detailed power consumption for TX2. As can be seen, the most power saving by \EP~comes from GPU processing, memory usage and WiFi data transmission. This is again
attributed to the fewer cameras performing 2D pose estimation (which involves heavy GPU computation) and sending the results to the edge server (via WiFi). We note that the CPU power consumption does not change much between SA and \EP. This is because in our current design of \EP, even if a camera is in the PSM, it still continuously captures video frames (or load the frames from the dataset), which requires CPU processing. Nevertheless, we can easily extend \EP~ to allow cameras to enter a more aggressive PSM where the unselected camera does not even capture/load the video frames. With this aggressive PSM mode, our experiment shows that the CPU power consumption can be further reduced by $17.64\%$ compared with SA.

\textbf{Demo using Live Videos.}
Figure \ref{fig:liveVideoResult} demonstrates an example multi-human 3D pose estimation result on live videos captured by the Logitech C270 HD Webcams connected to the NVIDIA devices in a lab environment. The estimated 3D poses are also re-projected onto the 2D images to offer a visual evaluation. The re-projected 2D skeletons closely fit the actual persons in the images, indicating that 3D and 2D poses are reliably estimated, even with occlusions by other people.

\begin{figure}[tt]
	\centering  
	\subfigure[]{
		\label{fig:iouOcclusionPrediction}
		\includegraphics[width=0.47\linewidth]{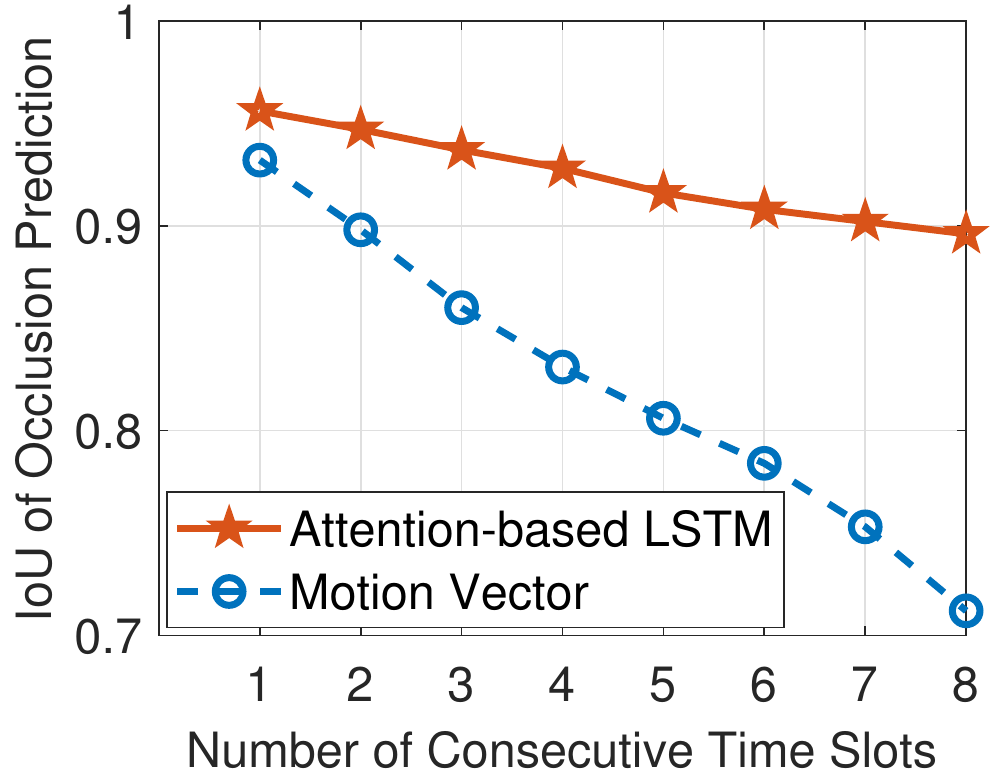}}
	\subfigure[]{
		\label{fig:pcpLatencyDiffTau}
		\includegraphics[width=0.47\linewidth]{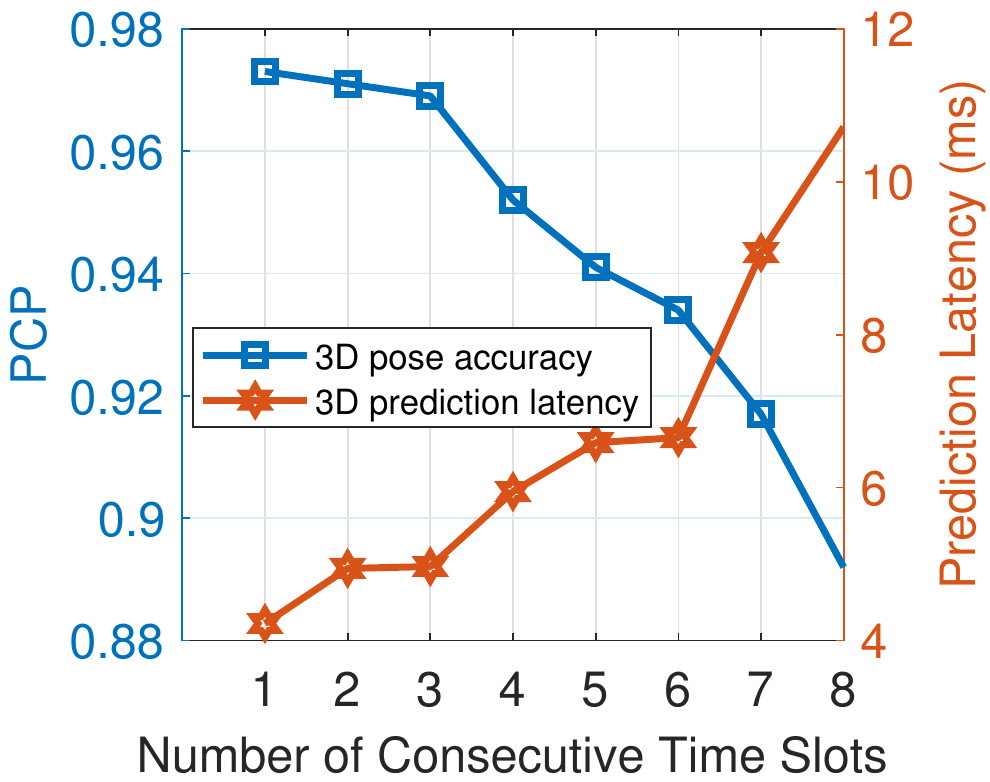}}
	\vspace{-15 pt}
	\caption{Performance of the occlusion prediction. (a) Occlusion prediction accuracy in IoU v.s. number of consecutive time slots $\tau$. (b) 3D pose estimation accuracy in PCP and 3D prediction latency vs. time advance $\tau$.}
	\label{fig:occlusionPrediction}
\end{figure}

\begin{figure}[tt]
	\centering
	\begin{minipage}[t]{0.48\linewidth}
		\includegraphics[width=\textwidth]{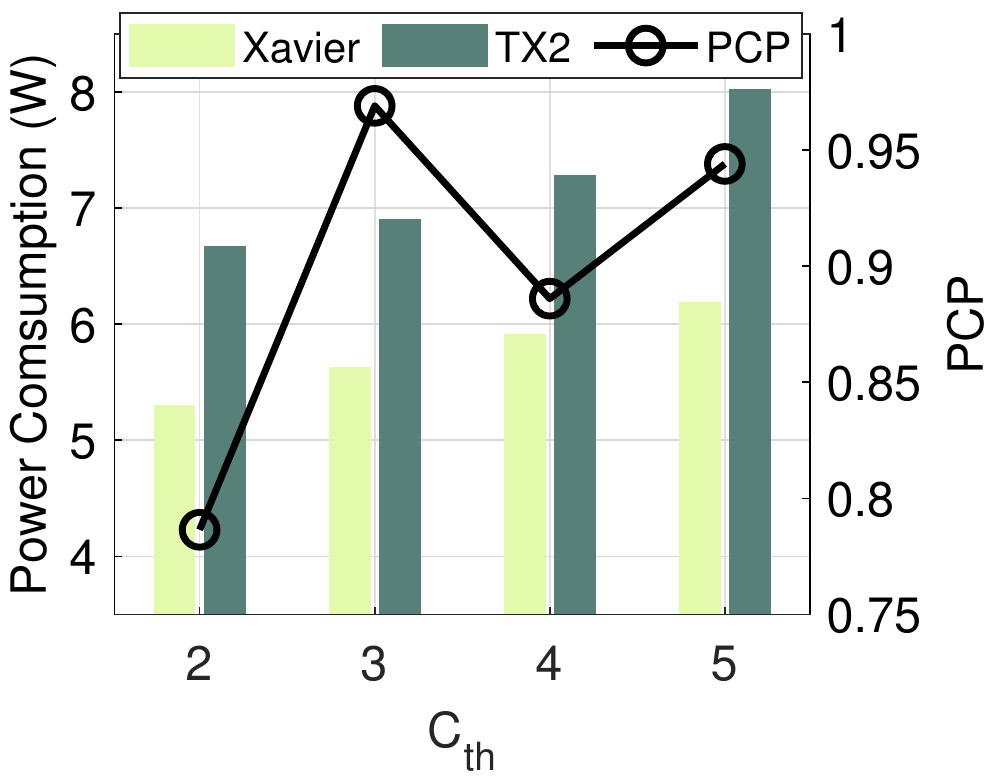}
		\vspace{-20 pt}
		\caption{Impact of the selected number of cameras $C$. (The test dataset is Shelf.)}
		\label{fig:cameraThresholdImpact}
	\end{minipage}
	\hspace{0.01\linewidth}
	\begin{minipage}[t]{0.48\linewidth}
		\includegraphics[width=\textwidth]{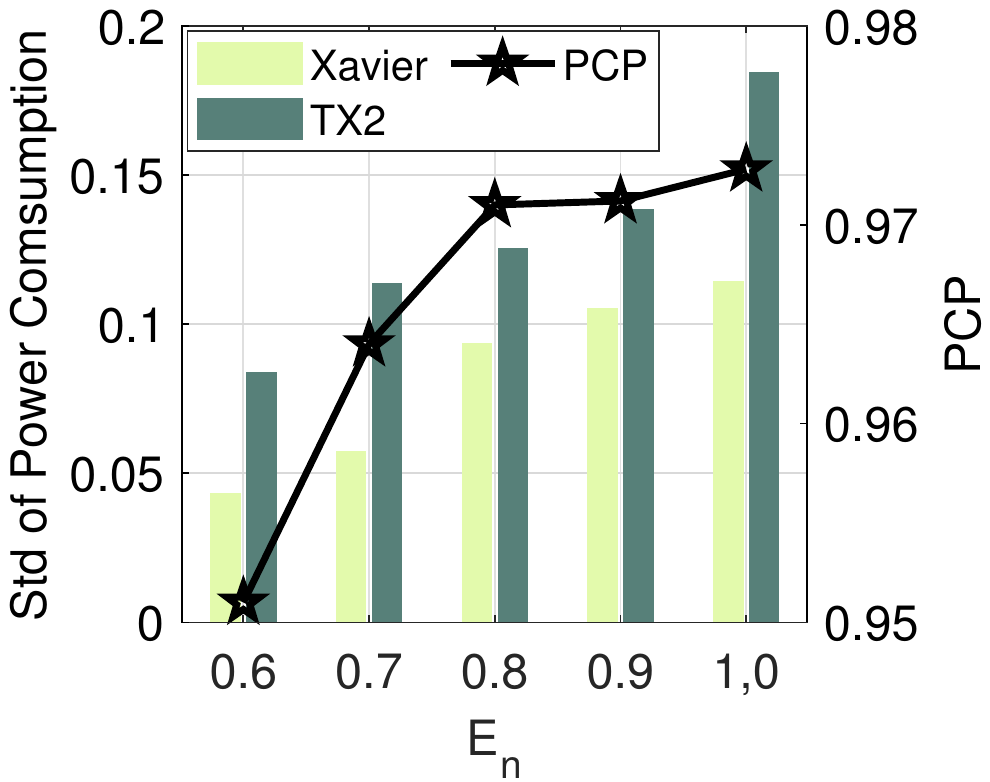}
		\vspace{-20 pt}
		\caption{Impact of the energy constraint $E_n$. (The test dataset is Shelf.)}
		\label{fig:imapctEnergyConstraint}
	\end{minipage}
\end{figure}

\subsection{Performance of Occlusion Prediction}
\label{sec:performanceOfOcclusionPrediction}

In this set of experiments, we focus on the occlusion prediction accuracy as it plays a key role in the camera selection. To this end, we consider the IoU between the predicted 2D bounding boxes (which are obtained by projecting the predicted 3D poses on the camera views) and the actual estimated 2D bounding boxes (which are obtained by 2D pose estimation on the image). Figure \ref{fig:iouOcclusionPrediction} shows the IoU values for \EP~ and \EP-MV. In \EP-MV, we use the 3D poses in two consecutive previous time slots to calculate the 3D motion vector for each human joint, and then use it to predict the human pose in the future time slots. However, since \EP-MV only considers the current motion of the human joints, its occlusion prediction accuracy decreases dramatically with a larger time advance $\tau$. On the contrary, the adopted attention-based LSTM can track the fine-grained potential motion of the humans and hence, it achieves a higher occlusion prediction accuracy than the motion vector-based prediction and is less sensitive to the time advance $\tau$. In Figure \ref{fig:pcpLatencyDiffTau}, we show that a reduced occlusion prediction does translate to a lowered PCP of the 3D pose estimation due to selecting the suboptimal subset of cameras. However, again, the decrease is small and not very sensitive to the time advance $\tau$. Figure \ref{fig:pcpLatencyDiffTau} also shows that a larger $\tau$ increases the prediction latency since the workload of the attention-based LSTM is increased.

\subsection{Impact of Camera Number Threshold}
Although we decided to use 3 cameras in our testbed according to the results in the motivational experiments in Section \ref{sec:accuracy_3dfusion}, we also try other values for the number of selected cameras. Figure \ref{fig:cameraThresholdImpact} shows the results by varying $C$ from 2 to 5 and fixing $E_n = 0.8$. Note that SA is the special case where $C = 5$. In Figure \ref{fig:cameraThresholdImpact}, the overall system power consumption increases with $C$. This is intuitive as more cameras are selected to execute 2D pose estimation and send the results to the edge server, more energy consumption is incurred. However, there is no monotonic relationship between $C$ and the accuracy of 3D pose estimation, and setting $C = 3$ achieves the best result. Using more cameras in 3D pose estimation is not necessarily beneficial for accuracy if cameras with poor-quality views are included.

\subsection{Impact of Energy Constraint}
In this subsection, we show the impact of the energy constraint on the performance of \EP. In the camera scheduling problem \eqref{eq:objectiveFunc}, for a fixed $C$, there is a trade-off between the accuracy of 3D pose estimation and the power consumption of individual cameras. When the cameras' energy capacity $E_n, \forall n$ is large, it is easier for the camera scheduler to select the cameras with good-quality views without worrying much about creating processing hotspots. As a result, a higher 3D pose estimation accuracy can be achieved at a cost of less balanced power consumption patterns across the cameras. Conversely, when $E_n$ is small, the camera scheduler must be cautious about not creating processing hotspots and using up the battery of certain cameras too soon. Therefore, the power consumption is more balanced across the cameras but the 3D pose estimation accuracy is lower. Figure \ref{fig:imapctEnergyConstraint} illustrates this phenomenon by varying $E_n$ from 0.6 to 1.0 while fixing $C = 3$. We can see that both the standard deviation of power consumption across the cameras and the PCP increases with $E_n$. 



\section{Related Work}

\subsection{2D Pose Estimation} 
2D human pose estimation aims to automatically locate human body joints from images or videos. Deep neural network (DNN) based 2D human pose estimations have received significant attention recently due to its high accuracy \cite{cao2017realtime, cai2020learning, chen2018cascaded, cheng2020higherhrnet, sun2019deep}. However, these DNN based 2D human pose estimation methods cannot be directly applied to resource-constrained mobile devices because they use computation-intensive deep neural network models. To overcome this limitation, some researchers focus on accelerating DNN based 2D human pose estimations on mobile devices. MobiPose \cite{zhang2020mobipose} investigates human pose estimation on smartphone SoCs with three novel techniques: the motion-vector-based method for fast location
of the human poses across frames, a mobile-friendly DNN model with low latency and sufficient accuracy, and an efficient parallel DNN engine. Blazepose \cite{bazarevsky2020blazepose} and PoseNet \cite{to2021real} use light-weight DNN models to enable real-time 2D human pose estimations on mobile device. In this paper, \EP~ uses the DNN architecture in \cite{cao2017realtime} with significantly more lightweight
ResNet18 as the backbone feature extractor to obtain 2D human poses. Moverover, \EP~ employs TensorRT, a DNN accelerator, to efficiently run 
the 2D human pose estimation on memory-limited mobile devices at real time. Note that \EP~ is not limited to only the DNN architecture proposed in \cite{cao2017realtime}; any other on-device 2D human pose estimation methods can also be transferred to \EP.

\subsection{3D Pose Estimation} 
Depending on the number of input cameras, 3D human pose estimation methods are categorized into single-view-based methods \cite{sun2018integral,cheng2019occlusion,zhang2020inference,andriluka2010monocular,moon2019camera,guler2018densepose} and multi-view-based methods \cite{qiu2019cross, belagiannis20143d, belagiannis2014multiple, dong2019fast, chen2020cross, lin2021multi, simon2021, tanke2019iterative, remelli2020lightweight}. 

Due to the difficulty of multi-human 3D pose estimations in the monocular view, most of the single-view approaches are developed to construct a single person’s 3D
poses \cite{sun2018integral, cheng2019occlusion, zhang2020inference}, where the predicted pose does not
include absolute human joint 3D coordinates in the environment. Although much progress has been made for
multi-human 3D pose estimation in a single view \cite{andriluka2010monocular, moon2019camera, guler2018densepose}, there is still a large deviation when applying these techniques in different practical surveillance scenarios. In particular, the motion blur and occlusions occur in images.

To retrieve absolute location and handle occlusions, the studies of multi-view 3D pose estimation attract more attention recently. It can be applied in various applications,
such as sports analysis, video surveillance, animation, and
healthcare \cite{wang2021deep}. Most existing approaches \cite{qiu2019cross, remelli2020lightweight} for single person 3D pose estimation are developed based on the 3D
Pictorial Structure model (3DPS), which cannot be directly used in multi-person pose estimation due to the lack of cross-view matching of 2D poses. Most state-of-the-art multi-human 3D pose estimation methods \cite{belagiannis20143d, belagiannis2014multiple, dong2019fast, chen2020cross, lin2021multi, tanke2019iterative} match the 2D poses estimation results from cross-view cameras, and fuse the matched 2D poses into 3D human poses. However, these recent methods focus more on accuracy than efficiency, and they do not consider how to deploy these methods in a real-world system with real-time constraints. Recently, an edge-assisted 3D pose estimation system \cite{simon2021} uses distributed smart cameras for 2D pose estimation. Then, the 2D human poses are streamed over a network to a central edge server, where data association, cross-view matching, multi-view triangulation and post-processing are performed to fuse the 2D human poses into 3D human skeletons. However, this work uses all the available cameras to perform 3D pose estimation without considering the energy constraints of the system. 

\section{Conclusion}
In this paper, we proposed a novel energy-efficient edge-assisted multi-camera system for real-time multi-human 3D human pose estimation. We advocated an adaptive camera selection scheme, which is able to achieve the benefits of both extending the battery lifetime of the system and improving the 3D pose estimation accuracy, at the cost of a slightly larger number of cameras. As smart cameras become increasingly cheaper, our proposed system provides an affordable and flexible solution for real-time 3D human pose estimation without requiring expensive special devices.



\bibliographystyle{unsrt}

\bibliography{reference}
	  
\end{document}